%% file: main.tex
\documentclass[runningheads]{llncs}

\usepackage[mobile]{eccv}

\usepackage{eccvabbrv}

\usepackage{graphicx}
\usepackage{booktabs}

\usepackage[accsupp]{axessibility}  

\input{preamble}

\usepackage{hyperref}

\usepackage{orcidlink}

\usepackage{multirow}

\begin{document}

\title{RadarGen: Automotive Radar Point Cloud Generation from Cameras} 

\author{
Tomer Borreda$^{1}$ \quad Fangqiang Ding$^{1,2}$ \quad Sanja Fidler$^{3,4,5}$ \\
Shengyu Huang$^{3}$ \quad Or Litany$^{1,3}$ \\
}
\authorrunning{T. Borreda et al.}
\institute{{\small $^{1}$ Technion \quad $^{2}$ MIT \quad $^{3}$ NVIDIA \quad $^{4}$ University of Toronto \quad $^{5}$ Vector Institute
}
\\
\small \url{https://radargen.github.io/}}

\maketitle

\vspace{-0.2em}
\begin{center}
    \nolinenumbers 
    \centering
    \includegraphics[width=0.84\linewidth]{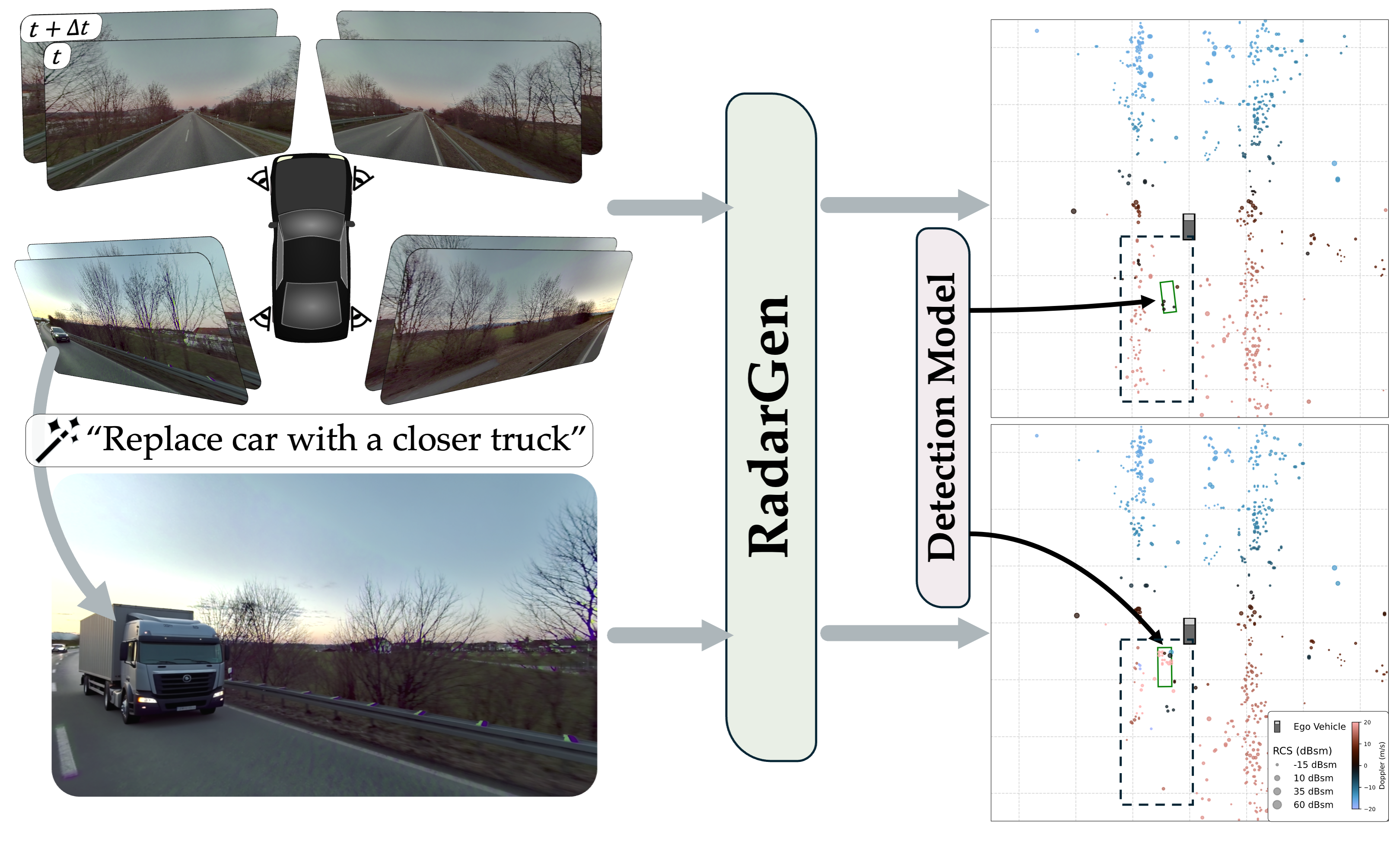}
\vspace{-0.4em}
    \captionof{figure}{\textbf{Controllable radar synthesis from vision.} (Top) Given multi-view camera images, \emph{RadarGen} generates realistic radar point clouds that align with real-world radar statistics and can be consumed by downstream perception models.
(Bottom) The generation is semantically consistent: modifying the input scene with an off-the-shelf image editing tool (e.g., replacing a distant car with a closer truck) updates the radar response, removing returns from newly occluded regions and reflecting the new object geometry.}
    \label{fig:teaser}
    \linenumbers 
\end{center}

\vspace{-0.7em}

\input{sections/00_abstract}

\input{sections/01_introduction}

\input{sections/02_related_work}

\input{sections/03_background}

\input{sections/04_method}

\input{sections/05_experiment}

\input{sections/06_conclusions}

\section*{Acknowledgments}
Or Litany acknowledges support from the Israel Science Foundation (grant 624/25) and the Azrieli Foundation Early Career Faculty Fellowship. The authors gratefully acknowledge this support. This research was supported by the Council for Higher Education in Israel under the Moonshot Project.

%
%
\bibliographystyle{splncs04}
\bibliography{main}

\clearpage
\title{Supplementary Materials for\\RadarGen: Automotive Radar Point Cloud Generation from Cameras} 
\titlerunning{RadarGen: Automotive Radar Point Cloud Generation from Cameras}
\authorrunning{T. Borreda et al.}
\author{}
\institute{}

\maketitle

\input{sections/07_suppl}

\end{document}

%% file: preamble.tex
\usepackage[T1]{fontenc}

\usepackage{comment}

\newcommand{\methodname}{\emph{RadarGen}}

\newcommand{\myparagraph}[1]{ \vspace{3pt}  \noindent {\bf #1}\,\,\,}

\definecolor{road}{RGB}{128,64,128}
\definecolor{sidewalk}{RGB}{244,35,232}
\definecolor{building}{RGB}{70,70,70}
\definecolor{vegetation}{RGB}{107,142,35}
\definecolor{car}{RGB}{0,0,142}
\definecolor{person}{RGB}{220,20,60}
\newcommand{\cbox}[1]{\textcolor{#1}{\rule{1.2ex}{1.2ex}}}

\usepackage{wrapfig}

%% file: sections/00_abstract.tex
\begin{abstract}
We present RadarGen, a diffusion model for synthesizing realistic automotive radar point clouds from multi-view camera imagery.
RadarGen adapts efficient image-latent diffusion to the radar domain by representing radar measurements in bird’s-eye-view form that encodes spatial structure together with radar cross section (RCS) and Doppler attributes.
A lightweight recovery step reconstructs point clouds from the generated maps.
To better align generation with the visual scene, RadarGen incorporates BEV-aligned depth, semantic, and motion cues extracted from pretrained foundation models, which guide the stochastic generation process toward physically plausible radar patterns.
Conditioning on images makes the approach broadly compatible, in principle, with existing visual datasets and simulation frameworks, offering a scalable direction for multimodal generative simulation.
Evaluations on large-scale driving data show that RadarGen captures characteristic radar measurement distributions and reduces the gap to perception models trained on real data, marking a step toward unified generative simulation across sensing modalities.
\end{abstract}

%% file: sections/01_introduction.tex
\section{Introduction}
Recent advances in neural and generative simulation have made it increasingly practical to synthesize photorealistic data at scale for autonomous driving.
By reconstructing real scenes with neural fields or generating entirely new ones using video diffusion models, these systems can produce diverse and controllable environments that closely mimic real sensor observations.
This capability enables large scale resimulation of traffic, lighting, and weather conditions without costly rerecording or manual setup~\cite{nvidia2025cosmosworldfoundationmodel,russell2025gaia,WeatherWeaver}.
Despite this rapid progress, most neural simulators remain limited to the visual domain, focusing on the generation of RGB imagery and video.
Recent efforts have begun extending these ideas to LiDAR, demonstrating controllable three-dimensional point cloud generation from camera inputs~\cite{ran2024towards,zyrianov2022learning,ren2025cosmos}.
Radar, however, remains an open frontier.
Although it is already ubiquitous in production vehicles, providing low cost, lightweight, and weather resilient perception, it has received far less attention within the generative modeling community.
This imbalance limits the fidelity of current neural simulators, which cannot reproduce radar’s distinctive sensing characteristics, including signal sparsity, radar cross section (RCS), and Doppler.

Generating radar data poses unique challenges.
Radar measurements exhibit strong stochasticity due to multipath reflections, interference, and material-dependent scattering that vary with scene geometry.
Operating at longer wavelengths, radar interacts with surfaces and internal structures beyond what cameras or LiDAR perceive, making it highly complementary yet difficult to model from vision alone.
A further challenge lies in the nature of available radar data.
In most large scale driving datasets, radar is provided only after proprietary signal processing that converts raw radio frequency waveforms into sparse point clouds with RCS and Doppler values.
This processing chain, which includes range-Doppler transforms, beamforming, and detection algorithms such as constant false alarm rate (CFAR), is closed and lossy, discarding phase and other fine grained signal information.
In practice, storing raw radar signals is extremely memory intensive, so even commercial survey vehicles often record only processed point clouds.
As a result, we opted for a point cloud representation as it remains the practical representation of radar data, providing diversity and scalability critical for generative model training.

To address these challenges, we propose \methodname{}, a generative framework for synthesizing automotive radar point clouds directly from camera imagery.
\methodname{} learns a distribution over radar observations conditioned on the visual scene, producing diverse and semantically consistent measurements rather than a single deterministic prediction, reflecting the inherent stochasticity of real radar signals.
Conditioning on images allows \methodname{} to leverage existing visual data and simulators, providing a scalable and modular way to enrich them with realistic radar so that radar-dependent perception systems can operate on camera-only and edited scenes, without re-recording radar.

A key design choice in \methodname{} is to build on SANA~\cite{xie2024sana}, an efficient image-latent diffusion model proven effective and scalable for image synthesis.
To the best of our knowledge, no existing generative model can reliably support scene level point cloud synthesis from multi-view images of real driving scenes, making an image diffusion backbone a natural and practical foundation.
SANA’s architecture supports conditioning on visual input while efficiently handling the large number of tokens introduced by the multi-channel radar representation and by the additional conditioning cues incorporated later in our framework, allowing \methodname{} to remain computationally efficient and deployable in large scale simulation settings.

To make radar data compatible with the image diffusion backbone, we express each radar point cloud as an image-like bird’s eye view (BEV) representation.
This representation encodes spatial structure together with radar cross section (RCS) and Doppler attributes, enabling all radar channels to share a unified latent space and allowing us to reuse SANA’s pretrained autoencoder without modification.
A lightweight smoothing and recovery step preserves geometric accuracy while ensuring stable latent encoding.

Learning radar purely from images is inherently difficult, as it requires reasoning about depth, semantics, and motion.
To avoid forcing the denoiser to learn these cues from scratch, \methodname{} leverages pretrained foundation models that provide dense visual priors.
Depth, semantic, and motion cues extracted from each camera view are projected into bird’s eye view to align with the radar representation and are fused within the diffusion model. 

Since no established benchmark exists for radar point cloud generation, we introduce a dedicated suite of metrics that measures geometric fidelity, radar attribute fidelity, and distribution similarity, both for background and foreground points. We hope this benchmark will support future progress in radar generation.

We summarize our main contributions below.
\begin{itemize}
\setlength\itemsep{0em}
    \item We present \methodname{}, the first probabilistic diffusion framework to generate realistic automotive radar point clouds including location, RCS, and Doppler from multi-view camera inputs.
    \item We introduce a latent diffusion methodology that trains on a BEV image representation of sparse radar attributes, conditioned by BEV-aligned visual depth, semantic, and motion priors from foundation models.
    \item We establish comprehensive metrics for radar point clouds based on geometric fidelity,  radar attribute fidelity, and distribution similarity, validating our design through extensive ablations.
    \item We demonstrate that the generated radar data can be interpreted by detectors trained on real data and supports applications such as scene editing.
\end{itemize}

%% file: sections/02_related_work.tex
\section{Related Work}

\subsection{Physics-based radar simulation}
Physics-based simulators model the emission, propagation and reception of electromagnetic (EM) waves based on physical laws. To rigorously simulate time-domain EM propagation, some works directly solve Maxwell’s equations in the integral~\cite{capsoni1998physically} or differential~\cite{liu2019simulation, stowell2008investigation,giannakis2015realistic} form, offering high physical accuracy but are computation-intensive for real-time applications. For practical usage, other works approximate EM wave propagation based on geometric optics and ray-tracing~\cite{yun2015ray}. This technique has been widely adapted to enhance system-level fidelity~\cite{topak2011system, dudek2011millimeter, holder2019fourier,he2022channel}, improve scalability and real-time performance~\cite{hirsenkorn2017ray, thieling2020scalable, sligar2020machine}, and support specific modern applications~\cite{schussler2021realistic, gubelli2013ray, arnold2022maxray}. Commercial software also implement similar techniques~\cite{RemcomWaveFarer,dosovitskiy2017carla}. RadSimReal~\cite{bialer2024radsimreal} utilized \textit{standard} 3D reflection signals to lessen dependence on radar-specific internals in ray-tracing simulators.
Distinct from ray-tracing, graphics-based simulators~\cite{ouza2017simple, stetco2020radar, schoffmann2021virtual} exploit the rasterization pipeline of graphics engines as an efficient physics proxy, generating radar measurements from depth in real time.
However, the aforementioned physical simulators rely on manually created assets, demanding substantial engineering effort to cover long-tail conditions.

\subsection{Data-driven radar simulation}
The heavy engineering efforts of physics-based radar simulators have motivated data-driven alternatives. One line adapts NeRF~\cite{mildenhall2021nerf, lei2024sar,borts2024radar,huang2024dart,rafidashti2025neuradar} or 3D Gaussian Splatting~\cite{kerbl20233d, kung2025radarsplat,li2025sar} to radar, learning scene-specific representations for novel-view rendering. These reconstructions, however, require multi-view radar captures per scene and transfer poorly to unseen scenes. In contrast, generative radar simulation methods synthesize measurements directly from conditioning inputs, enabling controllable generation for novel, unobserved environments. Prior works employ GANs~\cite{goodfellow2020generative} and VAEs~\cite{kingma2019introduction} conditioned on object distance~\cite{fidelis2023generation}, scene layouts~\cite{de2020generating,wheeler2017deep}, elevation maps~\cite{weston2021there}, or LiDAR~\cite{li2025sar}, but they ignore images as conditioning.

Two recent works~\cite{chen2023rf, deng2023midas} use visual conditioning for radar generation, but
focus on human-centric scenarios and depend on explicit physical modeling of signal-scene interactions, which limits scalability. For autonomous driving, Xiao et al.\cite{xiao2025simulate} synthesize radar cubes with a U-Net conditioned on camera/LiDAR plus waveform-parameter embeddings, while Rangaraj et al.~\cite{10896467} generate range-azimuth maps using an autoencoder conditioned on depth and segmentation. Yet these methods target radar raw data that is unavailable for large-scale training~\cite{ding2024radarocc}. Closer to our setting, Song et al.\cite{song2025simulating} and  Alkanat et al.~\cite{10447724} produce radar point clouds from LiDAR/RGB with convolutional networks, but their deterministic mappings under-model radar stochasticity and also do not exploit large pretrained foundation models for comprehensive scene encoding.

\subsection{Generative point cloud models}
Generative models specific to {radar} point clouds remain scarce in literature, so we review general point cloud generative frameworks as references. Many works designed unconditional models~\cite{zhou20213d,luo2021diffusion, zamorski2020adversarial,achlioptas2018learning,mo2019structurenet,yang2019pointflow,cai2020learning,vahdat2022lion} for point cloud completion and generation or text-conditioned~\cite{nichol2022point,wu2023sketch} generative frameworks, which rely on large priors and lack spatial grounding and controllability. In contrast, some recent works~\cite{melas2023pc2,lee2025rgb2point,tyszkiewicz2023gecco} learn to generate point clouds conditioned on RGB images. However, these works are object-centric, only generating point clouds for object shapes. Instead, some methods are proposed to tackle scene-level generation of LiDAR point clouds with generative techniques such as GAN~\cite{caccia2019deep,sallab2019lidar}, VQ-VAE~\cite{xiong2023learning} and diffusion models~\cite{zyrianov2022learning,ran2024towards,nakashima2024lidar,hu2024rangeldm,wu2024text2lidar}. While effective, these methods usually lack image-based conditional design and work on dense LiDAR range images, which cannot transfer to sparse and non-uniformly sampled radar point clouds~\cite{ding2022self,ding2023hidden,pan2024ratrack}.

%% file: sections/03_background.tex
\section{Preliminaries}
\label{sec:preliminaries}

We start by reviewing the challenges in generating radar point clouds (\cref{sec:preliminaries-radar-challenges}), and how we can benefit from efficient diffusion models (\cref{sec:preliminaries-efficient-dm}). This sets the stage for our proposed radar point cloud generation model (\cref{sec:method}).

\input{figures/fig_lidar_vs_radar}

\subsection{Challenges in radar point cloud generation}
\label{sec:preliminaries-radar-challenges}
The key challenge in generative modeling of radar point clouds lies in their unique data characteristics. Radar point clouds are sparse, unordered 3D point sets with highly non-uniform sampling. Unlike LiDAR, whose returns are uniform along angular dimensions and can be reshaped into dense range images, radar detections arise from peak-based target extraction (e.g., CFAR~\cite{scharf1991statistical}), and cannot form dense, grid-aligned measurements. See \cref{fig:lidar-vs-radar} and the Supplementary Material for more information. Further, each radar point carries sensor-specific attributes beyond 3D position—Radar Cross-Section (RCS), a proxy for reflectivity, mainly affected by the object material, geometry and incident angles, and Doppler velocity, the measurement of relative radial velocity.
As a result of this sparse, non-grid structure, we cannot effectively represent radar data in a range-image format. We thus adopt a multi-image bird's-eye-view (BEV) representation as a scalable alternative that is well-suited to the data's sparse nature.

\subsection{Efficient diffusion models}
\label{sec:preliminaries-efficient-dm}
A significant challenge in generative modeling is the synthesis of large-scale, explicit 3D point clouds conditioned on images. As discussed in \cref{sec:preliminaries-radar-challenges} we use a multi-image BEV representation, which is inherently high-dimensional, posing a significant challenge for standard diffusion architectures.
To tackle this high-dimensional generation problem at a scene-level scale, we must leverage an efficient diffusion architecture.
Specifically, Latent Diffusion Models (LDMs)~\cite{rombach2021highresolution, ho2020denoising} which compress images into smaller representations. Standard LDMs are insufficient, as they typically use an autoencoder (AE) with an 8x downsampling factor and a diffusion backbone with $O(N^2)$ quadratic self-attention complexity. We therefore build upon SANA~\cite{xie2024sana}, a framework that achieves efficiency by employing an AE with 32x compression and replacing the costly self-attention with an $O(N)$ linear attention mechanism.

%% file: figures/fig_lidar_vs_radar.tex
\begin{figure*}[t]
  \centering
  \includegraphics[width=1\linewidth]{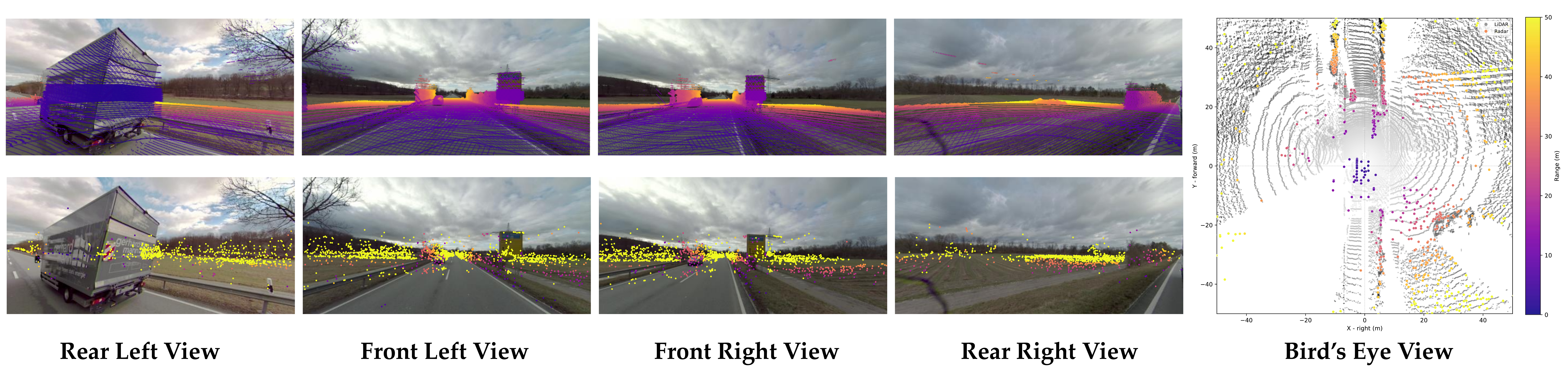}
  \vspace{-1.5em}
  \caption{
  Comparison of LiDAR and radar point clouds in camera and bird’s eye views. In the camera view, LiDAR is easier to interpret, whereas radar is more difficult due to its sparsity and lack of structure. In the bird’s eye view, radar points are easier to differentiate and interpret.
  }
\vspace{-0.5em}
  \label{fig:lidar-vs-radar}
\end{figure*}

%% file: sections/04_method.tex
\section{Method}
\label{sec:method}

\myparagraph{Problem statement.}
We address the task of surround-camera to radar point cloud generation.  
Given scene imagery captured by a rig of $N$ outward-facing cameras  
$\mathbf{I}^t = \{ I_1^t, \ldots, I_N^t \}$ at two consecutive timesteps $(t, t+\Delta t)$, with $\Delta t$ time elapsed between them,
together with known camera intrinsics and extrinsics,  
the goal is to generate a radar point cloud  
$\mathcal{P}^t = \{ (x_i, y_i, r_i, d_i) \}_{i=1}^{L}$  
representing the radar detections at time $t$ in the ego-vehicle coordinate frame.  
Each element describes the planar spatial coordinates $(x, y)$,  
radar cross section $r$ (RCS), and Doppler velocity $d$.  

\input{figures/fig_method}

\myparagraph{Method overview.}
\methodname{} consists of three main components (See \cref{fig:method} and \ref{fig:method-radar-maps}). During training:  
(1) the conversion of the \textit{sparse} radar point cloud $\mathcal{P}^t$ into a set of \textit{dense} BEV representations suitable for a latent image diffusion model (\cref{sec:method-radar-as-images}); and  
(2) conditioning the generator on geometric and semantic cues derived from pretrained vision models to enhance structural and semantic understanding (\cref{sec:method-conditional-generation}). 
With these we learn the conditional distribution 
\(
    p_\theta(\mathcal{P}^t \mid \mathbf{I}^t, \mathbf{I}^{t+\Delta t}),
\)
which models the stochastic mapping from multi-view imagery to the radar point cloud at time $t$.
At inference:  
(3) the recovery of the final sparse radar point cloud from the generated dense BEV predictions (\cref{sec:pcl-recovery-method}).

\input{figures/fig_method_radar_maps}

\subsection{Representing radar as images}
\label{sec:method-radar-as-images}
Our goal is to train a diffusion denoiser that operates in the latent space of a pretrained autoencoder.
To keep the autoencoder intact, the radar observations must therefore be represented as image-like signals that fit naturally within its latent distribution.  
However, radar measurements are sparse point clouds, far from the dense appearance statistics of natural images.  
We address this gap by transforming each radar point cloud into a set of dense, two-dimensional BEV maps that preserve radar-specific attributes while remaining compatible with the image-based AE. \cref{fig:method-radar-maps} illustrates the following steps.

Each radar point cloud is first projected onto the BEV plane by discarding elevation, which carries little discriminative information due to radar’s limited vertical resolution in automotive setups.
The projected points are then rasterized to form a BEV point map in which detections correspond to occupied pixels.

We construct three single-channel BEV maps: a Point Density Map ($M_p$), an RCS Map ($M_r$), and a Doppler Map ($M_d$). The Point Density Map $M_p$ is obtained by convolving the sparse BEV point map $\mathcal{P}_{xy}$ with a Gaussian kernel $K_\sigma$ of variance $\sigma$, i.e., $M_p = K_\sigma * \mathcal{P}_{xy}$, producing a smooth estimate of radar return density. Increasing $\sigma$ yields smoother maps with higher reconstruction fidelity under an AE, but also complicates point cloud recovery (see \cref{sec:pcl-recovery-method}).
For the RCS and Doppler maps, $M_r$ and $M_d$, each pixel inherits the attribute value (RCS or Doppler) of the nearest radar detection, yielding a piecewise-constant map defined by the Voronoi tessellation of the detections.

To ensure compatibility with the autoencoder’s RGB input space, each BEV map is replicated across three channels. We then encode each of these images independently, obtaining the latent representations $z_p$, $z_r$, and $z_d$ that serve as the supervision targets for training the diffusion denoiser.

Additional preprocessing details, including region-of-interest cropping and grid resolution, are provided in \cref{implementation-details}.

\subsection{Conditional radar generation}
\label{sec:method-conditional-generation}
To simplify the conditional radar generation task, we employ pretrained vision foundation models to extract relevant factors from conditional inputs and project them into BEV representations which are semantically rich and spatially aligned with the radar BEV targets.

Specifically, we use a depth estimation network \cite{piccinelli2025unidepthv2} to provide geometric cues, a semantic segmentation network \cite{cheng2021mask2former} to supply categorical context,  
and an optical flow model \cite{zhang2025ufm} to infer per-pixel motion.  
These cues are projected and fused into a unified bird’s-eye-view scene representation that serves as the conditioning input to the radar generator.

\myparagraph{BEV scene conditioning.}
The conditioning representation, denoted $\mathbf{c}$, comprises three BEV maps: an Appearance map, a Semantic map, and a Radial Velocity map (see \cref{fig:method}).  
We first project the $N$ outward-facing camera images $\mathbf{I}^t = \{I_1^t, \ldots, I_N^t\}$ into BEV using predicted metric depth~\cite{piccinelli2025unidepthv2}.
For each image $I_k^t$, a depth estimation model provides a dense point map $P_{I_k^t}$, which is then transformed into the ego-vehicle coordinate frame using the known camera intrinsics and extrinsics.  
All point maps are merged and rasterized onto a common BEV grid to form the geometric backbone of the condition.

The points that construct the \textit{Appearance} and \textit{Semantic maps} obtain their color from the original images and segmented images, respectively.
Using color-coded semantics, rather than one-hot encodings, ensures that the conditioning maps retain image-like statistics compatible with pretrained image encoders used later in our pipeline.

The \textit{Radial Velocity map} provides motion cues analogous to Doppler velocity. To construct it, we first estimate optical flow between consecutive frames $(I^t, I^{t+\Delta t})$ using a pretrained flow network~\cite{zhang2025ufm}.
Let $\mathbf{f}(x)$ denote the flow vector at pixel $x$ in $I^t$.
Using the predicted depth at time $t$ and $t+\Delta t$, we backproject $x$ in $I^t$ and its corresponding pixel $x + \mathbf{f}(x)$ in $I^{t+\Delta t}$ to 3D points $p^t(x)$ and $p^{t+\Delta t}(x)$ in the ego-vehicle frame.
We then approximate the 3D velocity as
\(
v(x) \approx \frac{p^{t+\Delta t}(x) - p^{t}(x)}{\Delta t},
\)
and retain only its component along the radial direction from the ego-vehicle to obtain a Doppler-like value.
These radial velocities are rasterized and aggregated into a BEV grid, yielding the Radial Velocity conditioning map.

This BEV scene representation provides pixel-aligned conditioning for the radar diffusion model, ensuring that geometric, semantic, and dynamic cues are spatially consistent with the generated radar BEV maps.

\myparagraph{Conditional radar maps denoising.}
With the radar maps represented in latent space, the learning objective becomes to model the distribution of plausible radar latents conditioned on the visual scene.  
We therefore train a conditional diffusion denoiser that learns $p(z_p, z_r, z_d \mid \mathbf{c})$, 
where $\mathbf{c}$ denotes the BEV conditioning maps.
Because both the conditioning and target latents are defined in a spatially aligned BEV coordinate frame,  
the conditioning tensors can be concatenated directly along the channel dimension \cite{ke2024repurposing}, providing the model with a compact yet expressive geometric prior.  
This formulation allows the network to generate diverse yet physically consistent radar realizations aligned with the observed camera views.

The denoiser $\boldsymbol{\epsilon}_\theta$ is implemented as a Diffusion Transformer (DiT)~\cite{peebles2023scalable} that operates jointly on the latent representations of the three radar maps—density, RCS, and Doppler.  
During each denoising step, the latents for these maps are concatenated into a single token sequence and processed through shared self-attention, enabling the network to capture correlations across radar modalities.  
Modality-specific identifiers, defined as learnable embeddings $m_i$ \cite{he2025unirelight} concatenated feature-wise to each radar map $i \in \{p,r,d\}$, guide the transformer to preserve the distinct statistics of each channel while still sharing information between them.  
At inference, Gaussian noise is iteratively denoised under BEV conditioning to produce the latent radar maps $\{z_p', z_r', z_d'\}$,  
which are then decoded into the final BEV radar images by the pretrained decoder $\mathcal{D}$.

\subsection{Recovering the sparse radar point cloud}
\label{sec:pcl-recovery-method}
As described in \cref{sec:method-radar-as-images}, the point density map is modeled as the convolution of a sparse point map with a Gaussian kernel,  
$M_p = K_\sigma * \mathcal{P}_{xy}$.  
The generative model therefore produces a blurred density estimate $M_p'$, from which the underlying discrete point locations must be recovered.

We formulate recovery as an explicit deconvolution problem, taking advantage of the fact that the blurring kernel $K_\sigma$ is known and fixed, since it was used to generate the training targets.
This knowledge of the forward blurring process enables us to pose a well-defined inverse problem for recovery, yielding a principled and controllable reconstruction of the sparse radar detections.

We solve for the sparse point map $\mathcal{P}_{xy}'$ via an L1-regularized, non-negative deconvolution (LASSO)~\cite{tibshirani1996regression}:
\begin{equation}
\min_{\mathcal{P}_{xy} \ge 0} \frac{1}{2}\|K_\sigma * \mathcal{P}_{xy} - M_p'\|_2^2 + \lambda \|\mathcal{P}_{xy}\|_1,
\label{eq:deconv}
\end{equation}
where $\lambda$ balances data fidelity and sparsity.  
We optimize this objective using an Iteratively Reweighted L1 (IRL1) \cite{daubechies2010iteratively} scheme with a FISTA~\cite{beck2009fast} solver, which provides fast convergence and stable recovery of sparse signals.  
The resulting $\mathcal{P}_{xy}'$ is thresholded to extract the final set of $L$ 2D coordinates $\{(x_i, y_i)\}_{i=1}^L$.  
For each recovered location, we retrieve the RCS and Doppler attributes from their corresponding map positions,  
producing the final reconstructed radar point cloud $\mathcal{P}' = \{(x_i, y_i, r_i, d_i)\}_{i=1}^L$.

\subsection{Implementation details}
\label{implementation-details}
Our diffusion model is a modified SANA DiT \cite{xie2024sana, peebles2023scalable} adapted for image conditioning, utilizing its pre-trained v1.1 autoencoder.  We use UnidepthV2 \cite{piccinelli2025unidepthv2} for metric depth estimation, Mask2Former \cite{cheng2021mask2former} (trained on Cityscapes \cite{cordts2016cityscapes}) for semantic segmentation, and UniFlow \cite{zhang2025ufm} for flow prediction. We trained the model for 2 days on 8 L40 (48GB) GPUs. During training, we drop each condition with a 10\% probability. We filter the radar point clouds to a $\pm 50\text{m}$ range and use a $512\times512$ grid.
Further implementation details are in the Supplementary Material.

%% file: figures/fig_method.tex
\begin{figure*}[t]
\begin{center}
    \includegraphics[width=\linewidth]{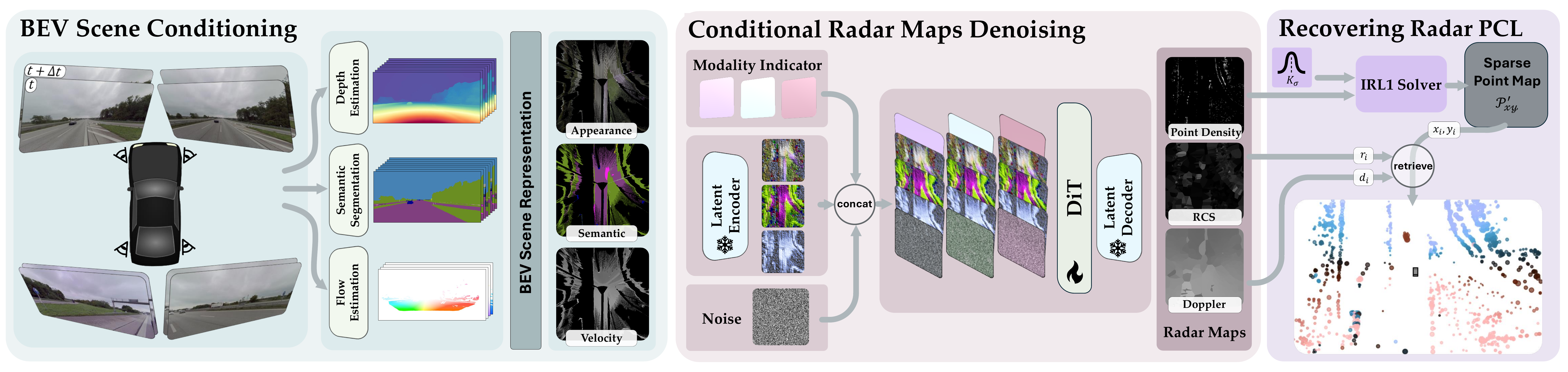}
\end{center}
    \vspace{-1em}
    \caption{\textbf{Overview of RadarGen.} (Left) Multi-view posed images at time $t$ and $t+\Delta t$ are fed through foundation models of metric depth estimation~\cite{piccinelli2025unidepthv2}, semantic segmentation~\cite{cheng2021mask2former}, and optical flow~\cite{zhang2025ufm}, enabling projection of the scene to BEV, encoding different information through color. (Middle) Encoded BEV representation is concatenated with a modality indicator specifying which map type to generate. During inference, the map is initialized as noise; during training, noise is added to the GT maps, and the Latent Encoder/Decoder are frozen while SANA's DiT~\cite{xie2024sana,peebles2023scalable} is fine-tuned. (Right) During inference, the generated Point Density Map is deconvolved using an IRL1 Solver. The resulting sparse map is used to retrieve the RCS and Doppler values at corresponding locations to yield the final generated radar point cloud. Point color represents Doppler and point size represents RCS.}
    \label{fig:method}
    
\vspace{-0.5em}
\end{figure*}

%% file: figures/fig_method_radar_maps.tex
\begin{figure}[t]
\begin{center}
    \includegraphics[width=0.7\linewidth]{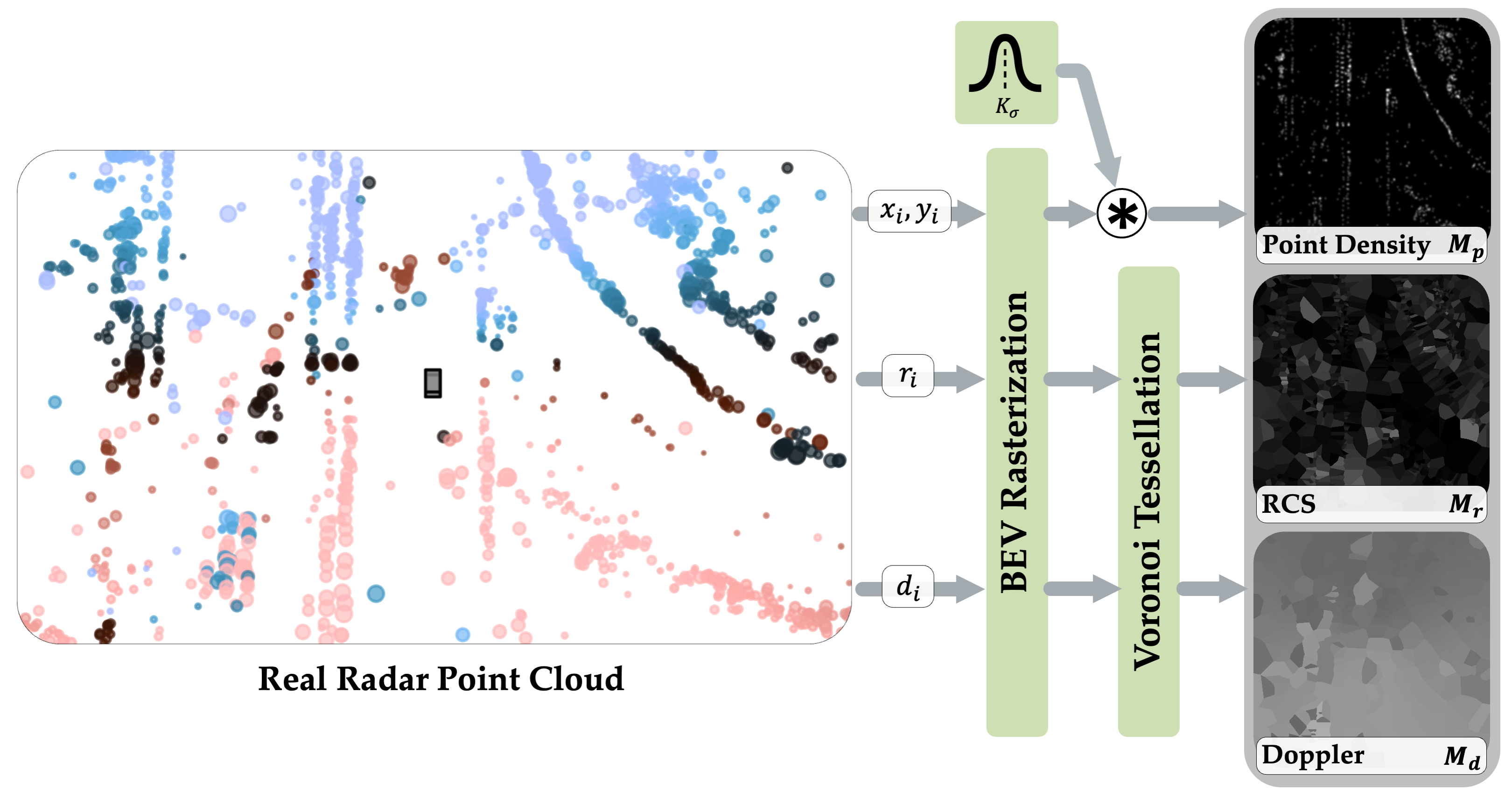}
\end{center}
    \vspace{-1em}
    \caption{\textbf{Overview of representing radar as images (\cref{sec:method-radar-as-images})}. Constructing radar maps from a radar point cloud requires first rasterizing each point to BEV. The point locations are then convolved with a Gaussian kernel $K_\sigma$ to produce the Point Density Map $M_p$. A Voronoi tessellation is also constructed, where each cell inherits the RCS and Doppler attributes from its corresponding point, producing the maps $M_r$ and $M_d$ respectively. Point color represents Doppler and point size represents RCS.}
    \label{fig:method-radar-maps}
\end{figure}

%% file: sections/05_experiment.tex
\input{tables/comparison}

\section{Experiments}
We start by describing the dataset, metrics, and baseline. In \cref{exp:pcl-gen-results}, we evaluate \methodname{} on radar point cloud generation from images. We then show how our proposed method can be used for scene editing in \cref{exp:app-scene-edit}. Finally, in \cref{exp:ablations} we analyze our design choices.

\myparagraph{Dataset.}
We evaluate on the MAN TruckScenes dataset~\cite{fent2024man} of driving scenes, which provides multi-view camera images and radar point cloud data with higher quality than other datasets, and includes rural and urban environments.
After excluding dark, uninformative scenes with near-zero visibility,
we use 431 clips for training and 49 for evaluation. While clips contain approximately 200 frames each, only 40 include bounding box annotations. We train on all frames in the training set and evaluate only on the annotated frames, which include bounding boxes for visible objects.
Evaluation on the nuScenes dataset \cite{caesar2020nuscenes} is available in the Supplementary Material.

\myparagraph{Evaluation metrics.}
As no standard benchmarks exist for our conditional radar generation task, we propose an evaluation framework. Our metrics assess three key aspects: geometric fidelity, radar attribute fidelity, and distribution similarity. Given the importance of foreground points for object detection, we calculate metrics for both the overall scene and points within annotated object bounding boxes. For geometric fidelity, we report \textit{Chamfer Distance} on location (CD Loc.) and the full vector of normalized location, RCS and Doppler (CD Full), Point Cloud \textit{IoU} at 1m (IoU@1m)~\cite{saunders2024baseboostdepth}, \textit{Density Similarity}, which measures the distance in number of points in a bounding box compared to the ground truth, and Bounding Box \textit{Hit Rate}. Radar attribute fidelity is assessed using \textit{Distance-Attribute} (DA) Recall, Precision, and F1-score, which assess the proximity of generated points with similar RCS and Doppler values to ground truth points. Finally, we evaluate distribution similarity using \textit{Maximum Mean Discrepancy} (MMD) for both entire point clouds and for points aggregated within object bounding boxes across scenes. Additional details on all metrics are available in the Supplementary Material.

\input{figures/fig_comparison}

\myparagraph{Baseline.}
We use the feedforward model RGB2Point \cite{lee2025rgb2point}, which is suitable for the task of predicting a point cloud from multi-view images. We extend the model output to include RCS and Doppler, and increase the number of parameters to 432M, which is comparable to our model with 592M parameters. See Supplementary Material for more implementation details about this baseline.

\subsection{Radar point cloud generation}
\label{exp:pcl-gen-results}

\myparagraph{Quantitative evaluation.} 
In \cref{tab:pcl_metrics} we compare \methodname{} to the baseline on the MAN TruckScenes dataset. \methodname{} broadly outperforms the baseline. A notable exception is CD Full in Entire Area, which is expected as the baseline model was trained using a similar loss objective.

\myparagraph{Qualitative evaluation.} In \cref{fig:comparison}, we visually compare our results with the baseline and ground truth on two examples. \emph{RadarGen's} generated point clouds closely match the ground truth in shape, distribution, and count, demonstrating a significant advantage over the baseline. Further visualizations and scene videos are available in the Supplementary Material.

\input{figures/fig_editing}

\myparagraph{Compatibility with perception models.}
We adapt and train the VoxelNeXt~\cite{chen2023voxelnext} detector on MAN TruckScenes radar data, which yields an NDS~\cite{caesar2020nuscenes} of $0.48$ on real data (50m range). We then evaluate this detector on generated radar point clouds (PCLs). On PCLs from \methodname{}, the detector scored an NDS of $0.30$. In contrast, the detector struggled to find valid objects in the baseline's PCLs, resulting in an NDS of nearly zero. Full results are available in the Supplementary Material.
Interestingly, while our generated radar achieves a high hit rate (0.66) within true bounding boxes, the overall detection quality still underperforms compared to real data. This could be attributed to the detector's tailoring to specific, intricate properties of the true data that our model may not fully capture. A detailed analysis of the subtle differences between real and generated radar data is beyond the scope of this paper.

Figure \ref{fig:teaser} visualizes detections on our generated PCL and demonstrates an augmentation scenario. In this example, a real vehicle is replaced with a generated truck; the detector successfully perceives the newly generated points as a truck.

\subsection{Scene editing}
\label{exp:app-scene-edit}
Our method supports radar point cloud augmentation by editing the input images using off-the-shelf image editing tools, such as ChronoEdit~\cite{wu2025chronoedit}. \cref{fig:teaser} demonstrates object replacement, while \cref{fig:obj-removal-insertion} shows examples of object removal and insertion. Notably, in the \cref{fig:teaser} replacement example (car to truck), the model correctly removes radar points from the area newly occluded by the truck, demonstrating that it properly handles occlusion changes.

\subsection{Ablation study and analysis}
\label{exp:ablations}

\input{tables/ablations}

\myparagraph{BEV conditioning ablations.}
We assess the contribution of each BEV condition by zeroing it out. Results are shown in \cref{tab:ablations}. Removing the segmentation map causes the most significant degradation. It worsens geometric fidelity and significantly increases the RCS MMD for both the Entire Area and per-object class in the Foreground. Ablating either the velocity map or the appearance map primarily degrades the Doppler MMD. Notably, the degradation from removing the appearance map, even with the segmentation map present, suggests the model leverages finer-grained appearance details to refine its understanding of object class and thus generate a more realistic motion profile. This confirms that all input channels are meaningfully fused to produce radar point clouds.

\myparagraph{Comparison to multi-view conditioning.}
We also compare against a model conditioned directly on multi-view (MV) camera images, which omits the BEV input. Instead, it uses images from $t$ and $t+\Delta t$ concatenated as input tokens with Plücker and modality embeddings. While this MV model achieves an improved MMD for radar attributes on the Entire Area, our BEV conditioned model yields better overall geometric fidelity, as detailed in \cref{tab:ablations}. Furthermore, the MV approach is computationally prohibitive, requiring over $3\times$ 
runtime; this model was trained for 9 days, compared to only 2 days for our BEV conditioned model.
While our BEV representation provides a more efficient and geometrically accurate solution, the results from the MV model highlight it as a valuable direction for future research.

\myparagraph{Radar PCL reconstruction.}
We ablate two factors: (a) the 2D Gaussian kernel bandwidth $\sigma$ for the Point Density Map, and (b) the PCL recovery method. We test $\sigma\in\{0.5, 1, 1.5, 2, 2.5, 3\}$ and four recovery methods (\texttt{random, peak, peak+random, deconv}) on the MAN TruckScenes \textit{mini-train} set, evaluating AE reconstruction error and the geometric fidelity of the recovered PCL.
As shown in Fig.~\ref{fig:ablation_sigma_recovery} (left), larger $\sigma$ values reduce AE reconstruction error, as smoother maps are easier to reconstruct. However, overly large $\sigma$ over-smooths the map structure, degrading downstream PCL recovery. Balancing this trade-off, we set $\sigma=2$. For the recovery method, \texttt{deconv} consistently yields the best PCL quality across all $\sigma$ values and for both ground-truth and AE-reconstructed maps, validating its ability to preserve the spatial distribution.

\input{figures/fig_recovery}

%% file: tables/comparison.tex
\begin{table*}[t]
\centering
\caption{
    \textbf{Quantitative evaluation.}
    \methodname{} broadly outperforms the baseline on geometric fidelity (CD, IoU, Density Similarity, Hit Rate), radar attribute fidelity (DA Recall, Precision, F1), and distribution similarity (MMD). Shown is the mean $\pm$ s.d., computed over all timestamps for the Entire Area and over all foreground objects for the Foreground.
}
\vspace{-0.5em}
\label{tab:pcl_metrics}
\resizebox{\textwidth}{!}{
\begin{tabular}{lccccccccc}
\toprule
\multirow{2}{*}{\textbf{Method}} & \multicolumn{9}{c}{\textbf{Entire Area}} \\
\cmidrule(lr){2-10}
& \multirow{1}{*}{\textbf{CD Loc. ($\downarrow$)}} & \multirow{1}{*}{\textbf{CD Full ($\downarrow$)}} & \multirow{1}{*}{\textbf{IoU@1m ($\uparrow$)}} & \multirow{1}{*}{\textbf{DA Recall ($\uparrow$)}} & \multirow{1}{*}{\textbf{DA Prec. ($\uparrow$)}} & \multirow{1}{*}{\textbf{DA F1 ($\uparrow$)}} & \multirow{1}{*}{\textbf{MMD Loc. ($\downarrow$)}} & \multirow{1}{*}{\textbf{MMD RCS ($\downarrow$)}} & \multirow{1}{*}{\textbf{MMD Doppler ($\downarrow$)}} \\
\midrule
Baseline & $1.84 \pm 0.48$ & $\mathbf{0.038 \pm 0.009}$ & $0.23 \pm 0.10$ & $0.15 \pm 0.10$ & $0.14 \pm 0.10$ & $0.14 \pm 0.09$ & $0.368 \pm 0.151$ & $0.36 \pm 0.25$ & $0.65 \pm 0.64$ \\
RadarGen & $\mathbf{1.68 \pm 0.39}$ & $0.040 \pm 0.008$ & $\mathbf{0.31 \pm 0.11}$ & $\mathbf{0.23 \pm 0.12}$ & $\mathbf{0.26 \pm 0.12}$ & $\mathbf{0.24 \pm 0.12}$ & $\mathbf{0.056 \pm 0.062}$ & $\mathbf{0.09 \pm 0.15}$ & $\mathbf{0.31 \pm 0.74}$ \\
\bottomrule
\end{tabular}
}

\resizebox{\textwidth}{!}{%
\begin{tabular}{lccccccccccccc}
\multirow{3}{*}{\textbf{}} & \multicolumn{13}{c}{\textbf{Foreground}} \\
\cmidrule(lr){2-14} 
& \multirow{2}{*}{\textbf{CD Loc. ($\downarrow$)}} & \multirow{2}{*}{\textbf{CD Full ($\downarrow$)}} & \multirow{2}{*}{\textbf{Density Sim. ($\uparrow$)}} & \multirow{2}{*}{\textbf{Hit Rate ($\uparrow$)}} & \multicolumn{3}{c}{\textbf{MMD Car ($\downarrow$)}} & \multicolumn{3}{c}{\textbf{MMD Truck ($\downarrow$)}} & \multicolumn{3}{c}{\textbf{MMD Trailer ($\downarrow$)}} \\ 
\cmidrule(lr){6-8} \cmidrule(lr){9-11} \cmidrule(lr){12-14}
& & & & & \textbf{Loc.} & \textbf{RCS} & \textbf{Doppler} & \textbf{Loc.} & \textbf{RCS} & \textbf{Doppler} & \textbf{Loc.} & \textbf{RCS} & \textbf{Doppler} \\
\midrule
Baseline & $1.32 \pm 0.79$ & $0.075 \pm 0.049$ & $0.35 \pm 0.43$ & $0.37$ & $\mathbf{0.035}$ & $0.753$ & $0.549$ & $0.167$ & $0.202$ & $0.485$ & $0.0459$ & $0.064$ & $0.607$ \\
RadarGen & $\mathbf{0.95 \pm 0.65}$ & $\mathbf{0.069 \pm 0.049}$ & $\mathbf{0.51 \pm 0.41}$ & $\mathbf{0.66}$ & $0.037$ & $\mathbf{0.006}$ & $\mathbf{0.014}$ & $\mathbf{0.024}$ & $\mathbf{0.031}$ & $\mathbf{0.060}$ & $\mathbf{0.0069}$ & $\mathbf{0.022}$ & $\mathbf{0.046}$ \\
\bottomrule
\end{tabular}
}
\vspace{-0.5em}
\end{table*}

%% file: figures/fig_comparison.tex
\begin{figure*}[t]
  \centering
  \includegraphics[width=1\linewidth]{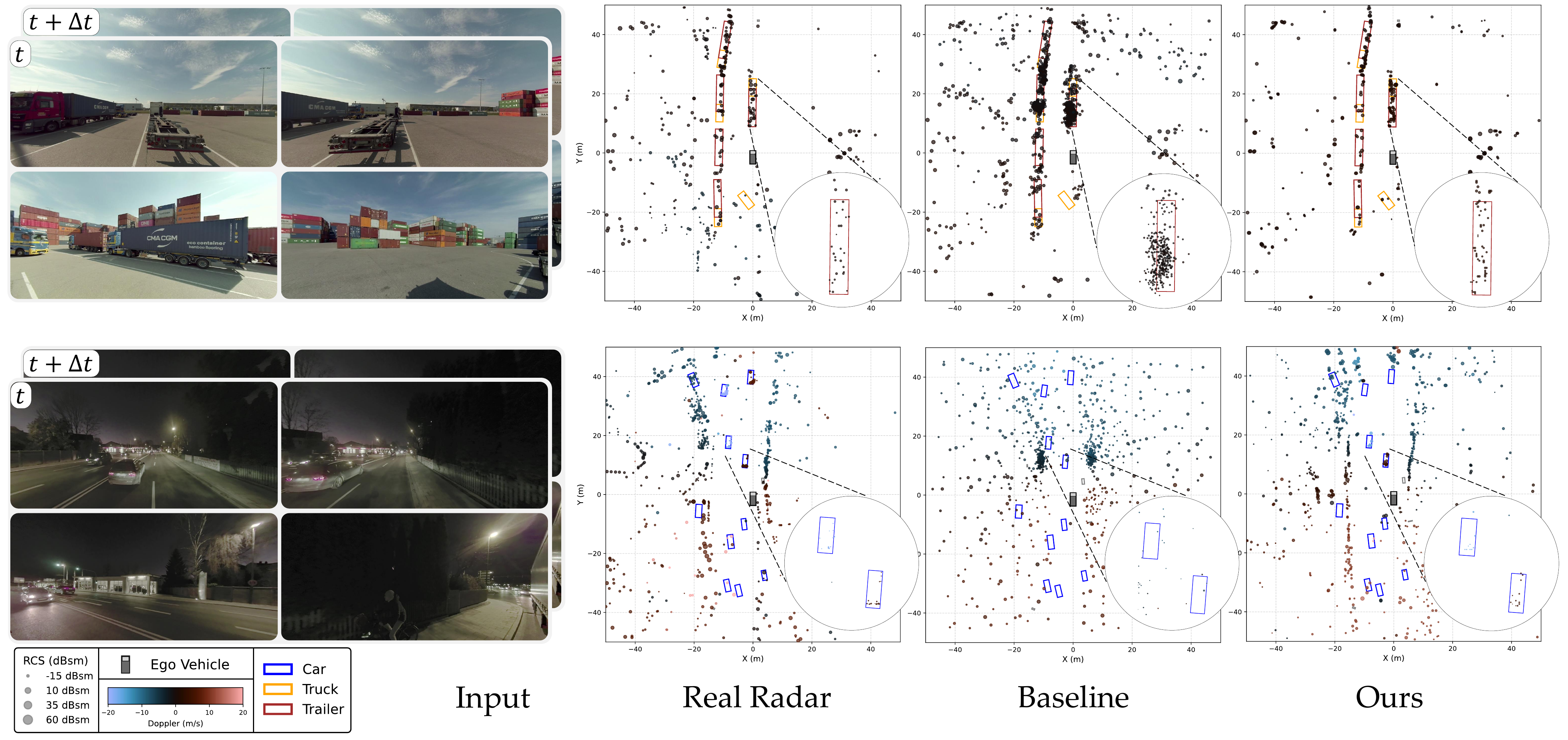}
  \vspace{-1.5em}
  \caption{\textbf{Qualitative results.} Our model generates point clouds with higher geometric and attribute fidelity to the ground truth compared to the baseline. \methodname{} uses inputs $t$ and $t+\Delta t$, while the baseline uses only $t$. Ground truth bounding boxes are highlighted in color.}
  \label{fig:comparison}
\end{figure*}

%% file: figures/fig_editing.tex
\begin{figure}[t]
  \centering
  \includegraphics[width=0.9\linewidth]{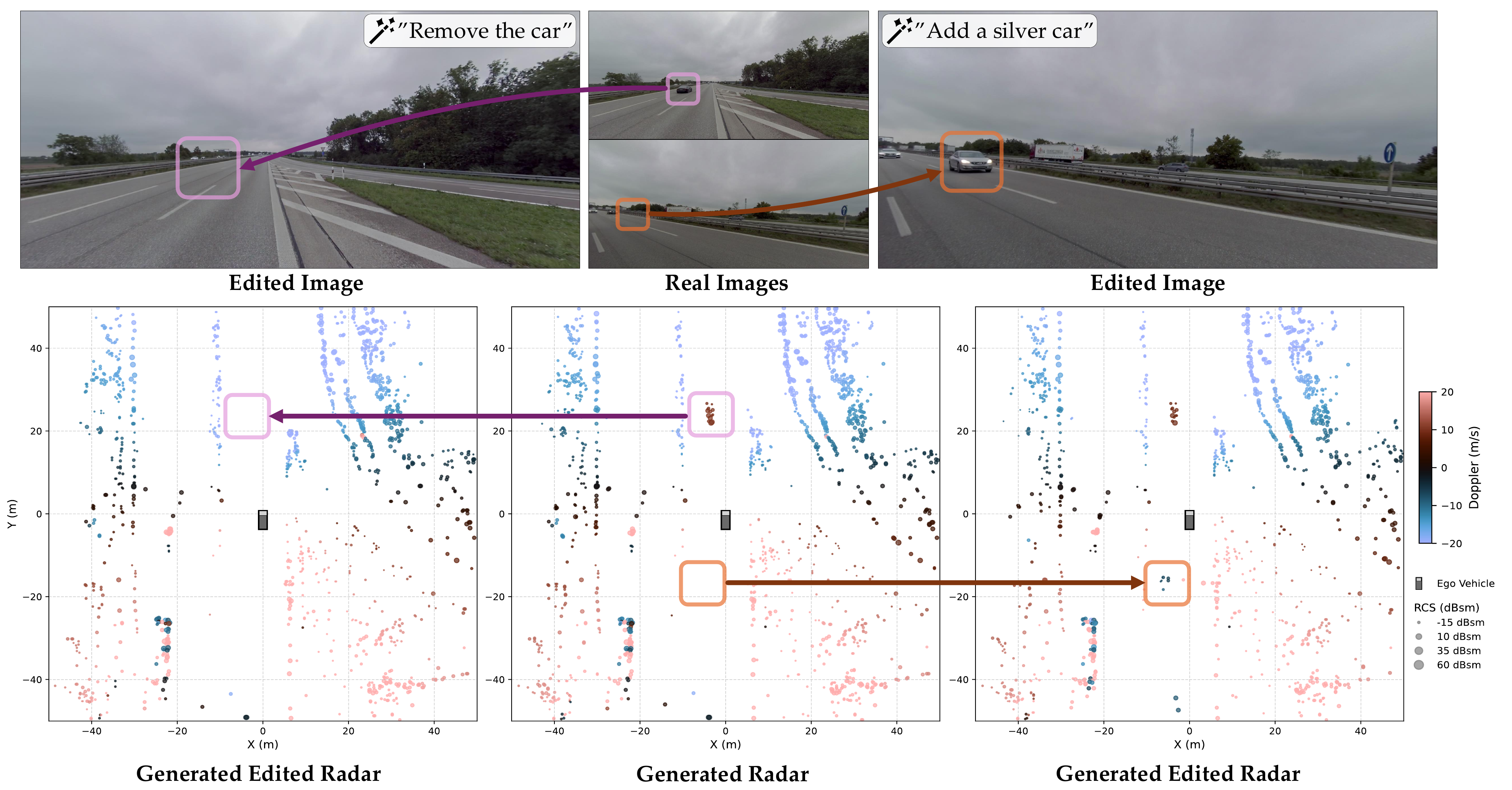}
  \caption{\textbf{Scene editing.} Modifying the input images using an off-the-shelf image editing tool updates the radar response, demonstrating object removal (left) and insertion (right).}
  \label{fig:obj-removal-insertion}
\end{figure}

%% file: tables/ablations.tex
\begin{table*}[t]
\centering
\vspace{-0.5em}
\caption{\textbf{Ablation Study.} We demonstrate the importance of each \methodname{} condition and compare against a model conditioned directly on multi-view (MV) camera images. Evaluation covers geometric fidelity (CD, IoU, Density Similarity, Hit Rate), radar attribute fidelity (DA Recall, Precision, F1), and distribution similarity (MMD). Shown is the mean $\pm$ s.d., computed over all timestamps for the Entire Area and over all foreground objects for the Foreground.
}
\vspace{-0.5em}
\label{tab:ablations}
\resizebox{\linewidth}{!}{%
\begin{tabular}{lccccccccc}
\toprule
\multirow{2}{*}{\textbf{Method}} & \multicolumn{9}{c}{\textbf{Entire Area}} \\
\cmidrule(lr){2-10}
& \multirow{1}{*}{\textbf{CD Loc. ($\downarrow$)}} & \multirow{1}{*}{\textbf{CD Full ($\downarrow$)}} & \multirow{1}{*}{\textbf{IoU @ 1m ($\uparrow$)}} & \multirow{1}{*}{\textbf{DA Recall ($\uparrow$)}} & \multirow{1}{*}{\textbf{DA Prec. ($\uparrow$)}} & \multirow{1}{*}{\textbf{DA F1 ($\uparrow$)}} & \multirow{1}{*}{\textbf{MMD Loc. ($\downarrow$)}} & \multirow{1}{*}{\textbf{MMD RCS ($\downarrow$)}} & \multirow{1}{*}{\textbf{MMD Doppler ($\downarrow$)}} \\
\midrule
MV Camera Cond. & $1.88 \pm 0.50$ & $0.041 \pm 0.009$ & $0.28 \pm 0.12$ & $0.26 \pm 0.13$ & $0.25 \pm 0.12$ & $0.25 \pm 0.11$ & $0.059 \pm 0.057$ & $0.06 \pm 0.09$ & $0.24 \pm 0.85$ \\

W/o Appearance map & $1.71 \pm 0.40$ & $0.040 \pm 0.008$ & $0.31 \pm 0.11$ & $0.23 \pm 0.12$ & $0.25 \pm 0.12$ & $0.23 \pm 0.12$ & $0.059 \pm 0.069$ & $0.09 \pm 0.16$ & $0.35 \pm 0.86$ \\
W/o Semantic map & $1.72 \pm 0.40$ & $0.041 \pm 0.010$ & $0.31 \pm 0.11$ & $0.22 \pm 0.12$ & $0.24 \pm 0.12$ & $0.23 \pm 0.12$ & $0.059 \pm 0.061$ & $0.12 \pm 0.26$ & $0.33 \pm 0.72$ \\
W/o Velocity map & $1.69 \pm 0.40$ & $0.040 \pm 0.008$ & $0.31 \pm 0.11$ & $0.23 \pm 0.12$ & $0.25 \pm 0.12$ & $0.23 \pm 0.12$ & $0.057 \pm 0.064$ & $0.09 \pm 0.16$ & $0.34 \pm 0.80$ \\

RadarGen & $1.68 \pm 0.39$ & $0.040 \pm 0.008$ & $0.31 \pm 0.11$ & $0.23 \pm 0.12$ & $0.26 \pm 0.12$ & $0.24 \pm 0.12$ & $0.056 \pm 0.062$ & $0.09 \pm 0.15$ & $0.31 \pm 0.74$ \\
\bottomrule
\end{tabular}
}


\resizebox{\linewidth}{!}{%
\begin{tabular}{lccccccccccccc} 
\multirow{3}{*}{\textbf{}} & \multicolumn{13}{c}{\textbf{Foreground}} \\ 
\cmidrule(lr){2-14} 
& \multirow{2}{*}{\textbf{CD Loc. ($\downarrow$)}} & \multirow{2}{*}{\textbf{CD Full ($\downarrow$)}} & \multirow{2}{*}{\textbf{Density Sim. ($\uparrow$)}} & \multirow{2}{*}{\textbf{Hit Rate ($\uparrow$)}} & \multicolumn{3}{c}{\textbf{MMD Car ($\downarrow$)}} & \multicolumn{3}{c}{\textbf{MMD Truck ($\downarrow$)}} & \multicolumn{3}{c}{\textbf{MMD Trailer ($\downarrow$)}} \\ 
\cmidrule(lr){6-8} \cmidrule(lr){9-11} \cmidrule(lr){12-14} 
& & & & & \textbf{Loc.} & \textbf{RCS} & \textbf{Doppler} & \textbf{Loc.} & \textbf{RCS} & \textbf{Doppler} & \textbf{Loc.} & \textbf{RCS} & \textbf{Doppler} \\ 
\midrule

MV Camera Cond. & $1.0 \pm 0.67$ & $0.069 \pm 0.050$ & $0.47 \pm 0.42$ & $0.56$ & $0.025$ & $0.018$ & $0.024$ & $0.017$ & $0.070$ & $0.073$ & $0.0029$ & $0.037$ & $0.040$ \\

W/o Appearance map & $0.95 \pm 0.64$ & $0.069 \pm 0.050$ & $0.51 \pm 0.41$ & $0.65$ & $0.034$ & $0.006$ & $0.019$ & $0.019$ & $0.019$ & $0.059$ & $0.0048$ & $0.026$ & $0.041$ \\
W/o Semantic map & $0.96 \pm 0.66$ & $0.070 \pm 0.050$ & $0.50 \pm 0.41$ & $0.64$ & $0.045$ & $0.010$ & $0.017$ & $0.023$ & $0.039$ & $0.078$ & $0.0072$ & $0.032$ & $0.061$ \\
W/o Velocity map & $0.95 \pm 0.65$ & $0.069 \pm 0.049$ & $0.51 \pm 0.41$ & $0.66$ & $0.033$ & $0.006$ & $0.019$ & $0.024$ & $0.018$ & $0.070$ & $0.0051$ & $0.024$ & $0.047$ \\

RadarGen & $0.95 \pm 0.65$ & $0.069 \pm 0.049$ & $0.51 \pm 0.41$ & $0.66$ & $0.037$ & $0.006$ & $0.014$ & $0.024$ & $0.031$ & $0.060$ & $0.0069$ & $0.022$ & $0.046$ \\
\bottomrule
\end{tabular}
}
\vspace{-1em}
\end{table*}

%% file: figures/fig_recovery.tex
\begin{figure}[t]
  \centering
  \includegraphics[width=0.7\linewidth]{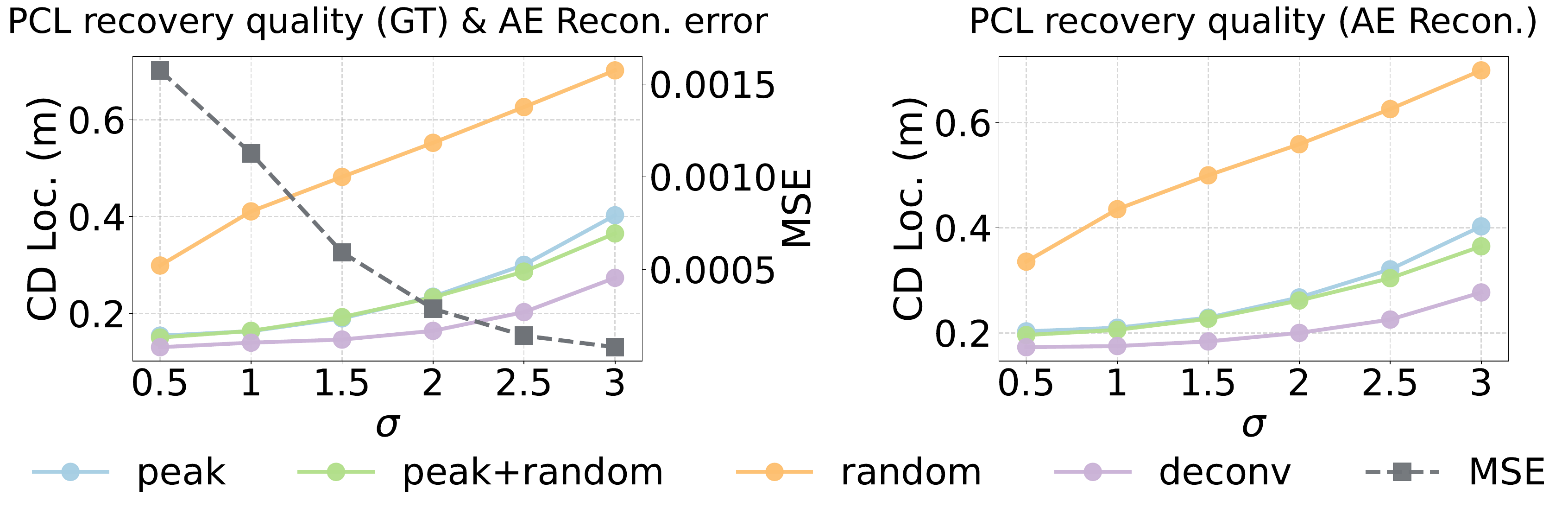}
  \caption{Impact of \(\sigma\) and recovery method on PCL reconstruction. \texttt{peak} selects local maxima; \texttt{random} samples pixels proportional to the probability map (without replacement); \texttt{peak+random} takes peaks first, then fills remaining points by probability sampling excluding the peaks. MSE is calculated between the input and output of the AE, whose range is [-1,1].}
  \label{fig:ablation_sigma_recovery}
\end{figure}

%% file: sections/06_conclusions.tex
\section{Conclusion and Future Work}
In this paper, we introduced \methodname{}, a probabilistic diffusion framework for generating realistic automotive radar point clouds from multi-view camera inputs. Our method leverages foundation models to create a unified BEV representation for conditioning, generates dense BEV radar maps, and uses a deconvolution solver to recover the final sparse point cloud. Experimentally, \methodname{} outperforms our proposed baseline on a comprehensive suite of geometric, attribute, and distribution metrics. We demonstrate its application in data augmentation via simple image editing and show promising results for downstream detectors. Future work will focus on extending our framework for video input, exploring text-based conditioning, and training on multiple datasets and radar configurations.

\textbf{Limitations.} Our method's performance is inherently tied to the capabilities of the upstream foundation models. It is thus limited in challenging scenarios where these models underperform, such as in low-light night scenes, with strong reflections, or during camera occlusion. Additionally, our model can generate points in areas not directly visible to the cameras. This behavior is a double-edged sword: while it is desirable for "filling in" occluded objects, it can also lead to uncontrolled hallucinations in these unseen regions. Finally, in this work we did not model explicit radar mechanisms, as doing so requires large-scale raw data and detailed radar specifications and internal mechanisms, which are often undisclosed. If large-scale raw data becomes available, modeling explicit radar properties would be a fruitful next step.

%% file: sections/07_suppl.tex
\appendix

\noindent This supplementary material includes evaluation metrics formulation, additional implementation details and additional experimental results.

\addtocontents{toc}{\protect\contentsline {section}{APPENDIXSTART}{}{}}
\newif\ifinappendix
\inappendixfalse

\begingroup
\makeatletter
\let\clearpage\relax 
\renewcommand{\contentsname}{} 

\setcounter{tocdepth}{2} 

\let\oldcontentsline\contentsline
\renewcommand{\contentsline}[4]{%
  \def\temp{#2}%
  \def\marker{APPENDIXSTART}%
  \ifx\temp\marker
    \inappendixtrue
  \else
    \ifinappendix
      \oldcontentsline{#1}{#2}{#3}{#4}%
    \fi
  \fi
}%

\vspace{1.5em}
\noindent{\large\bfseries Contents} 
\vspace{-2.5em}

\tableofcontents
\makeatother
\endgroup

\input{sections/07_supp_eval}

\section{Additional Details}

\subsection{Additional details on BEV representation}
\myparagraph{Radar maps.}
Before converting radar maps to latent representation we require an image that would serve as the encoder input. Maps $M_p,M_r,M_d$ possess values in the following ranges:
\begin{equation*}
\begin{array}{lcl}
M_p & \in & \left[-50\text{m}, 50\text{m}\right], \\
M_r & \in & \left[-20\text{dBsm}, 66\text{dBsm}\right], \\
M_p & \in & \left[-120\text{m/s}, 120\text{m/s}\right].
\end{array}
\end{equation*}
We use these values to normalize the maps to the range $[0,255]$.

\myparagraph{BEV scenes conditioning.}
When converting images to point maps and subsequently to a $512\times512$ BEV grid, we filter the point maps to reduce input noise. Specifically, we discard points corresponding to object edges and sky regions. We also remove points with a predicted height greater than 5m to filter out overhead structures such as bridges and trees. Finally, if multiple points occupy a single grid cell, we retain only the point with the maximum height. \cref{fig:supp_bev_condition} shows visualizations of the BEV conditioning maps.

\myparagraph{Explainability.}
A key advantage of BEV scene conditioning is interpretability; it clarifies exactly what information is available to the model, allowing us to anticipate and explain the output radar maps. This contrasts with models conditioned solely on camera images, where the model's internal awareness of object existence, classification, and location remains opaque.

\subsection{Additional details on training}
We train \methodname{} for 2 days on 8 L40 (48GB) GPUs, totaling 65k steps. We initialize the model using SANA's \cite{xie2024sana} pre-trained 600M-parameter weights at $512\times512$ resolution, utilizing SANA AE v1.1. We fine-tune the model using a batch size of 16 per GPU, seed 42, gradient accumulation of 2, bf16 mixed precision, and a learning rate of $10^{-4}$. Additionally, we apply conditioning dropout with a probability of 10\%, replacing the condition with a zero tensor. To improve training efficiency, we pre-compute and store the required BEV images during a preprocessing step.
On Man TruckScenes the time elapsed between consecutive camera timestaps, $\Delta t$, is approximately $0.1s$. On nuScenes this time may vary. In practice we compute the time between consecutive timestamps independently for each frame pair on each view.

\subsection{Additional details on inference}
During inference of our diffusion model we use 20 sampling steps, a null prompt, and no guidance scale. For reproducibility, the initial seed is set to 42.
During the sparse point cloud recovery, we deconvolve the radar maps via an IRL1 solver with FISTA. We utilize the following hyperparameters: $\lambda=0.0018$, 300 FISTA iterations, 5 IRL1 iterations, and a threshold of 0.1 for the sparse map $\mathcal{P}_{xy}'$.

\myparagraph{Inference time.}
The total inference time for a single timestep $t$ on a single L40 GPU is approximately 10.5 seconds. This decomposes into 9 seconds for BEV conditioning map creation, 1 second for diffusion inference, and 0.5 seconds for point cloud recovery.

\subsection{Additional details on the baseline model}
\label{supp:baseline-details}
We employ RGB2Point~\cite{lee2025rgb2point} as our multi-view image-to-point cloud baseline. This model utilizes a pre-trained ViT to extract features from an arbitrary number of input views and maps them to a point set. We adapt the architecture to predict RCS and Doppler attributes by increasing the point cloud dimension from 3 to 5. Additionally, we increase the number of attention heads to 8, the intermediate linear dimension to 2048, and the feedforward dimensionality to 4096. Input images are resized to $224\times224$ and normalized following the original method. We retain the Chamfer Distance (CD) loss, computed across all 5 channels. The output is a fixed set of 1024 points. The model was trained with a batch size of 8 and a learning rate of $10^{-5}$ on 8 L40 (48GB) GPUs for 18 hours.
We observed that the model failed to converge when using the normalized ``CD Full'' objective; therefore, we opted not to normalize the values and minimized the Chamfer Distance on the full unnormalized vector.

\subsection{Additional details on VoxelNeXt}
\label{supp:voxelnext-details}
We adapt VoxelNeXt \cite{chen2023voxelnext} for radar input, training for 77k steps (9 hours) on 8 L40 (48GB) GPUs. We use a batch size of 32 per GPU and apply data augmentations including flipping, rotation, translation, and scaling. The point cloud range is set to $[-50\text{m}, 50\text{m}]$.

\subsection{Additional details on direct multi-view conditioning}
To condition radar maps generation on multi-view camera images, we apply the same shared self-attention mechanism used for multi-map denoising. The denoising process involves 11 latent tensors: 3 for the radar maps, 4 for camera images at time $t$, and 4 for camera images at time $t+1$.
Each image is transformed to the model's required  $512\times512$ resolution by padding and resizing. Each image is then encoded to into a latent tensor and concatenated along the features dimension with its corresponding Plücker coordinates (adjusted for padding and relative to the ego vehicle at time $t$) and a unique modality indicator. Similarly, the radar map latents are concatenated with a zero tensor to match the feature dimensionality, along with a unique modality indicator.
We train this model for 65k steps on 8 L40 (48GB) GPUs, which takes 9 days, using the same hyperparameters as \methodname{}.

\section{Additional Results}

\subsection{Additional qualitative results}
We provide additional qualitative results for highway and rural environments (\cref{fig:supp_comparison_open_road}), as well as urban areas (\cref{fig:supp_comparison_urban}). We also illustrate the effect of different random seeds (\cref{fig:supp_multiple_seeds}). Finally, we include supplementary videos demonstrating scene generation.

\subsection{Additional radar PCL recovery results}
\cref{fig:supp_recovery} compares the Random, Peak, and Deconvolution sparse point cloud recovery methods on generated scenes. The Random sampling method treats the point density map as a probability distribution, resulting in an inconsistent point distribution. This causes gaps in the data; for instance, in the third row, the moving car is not captured, while other areas exhibit excessive density. The Peak method, which selects the local maximum within a $3\times 3$ window, results in a sparse point cloud that captures all relevant regions but lacks sufficient density. The Deconvolution method combines the strengths of both approaches, covering all relevant regions while ensuring sufficient density where required.

\subsection{Additional information on limitations}
\cref{fig:supp_limitation} illustrates a low-light night scene where the underlying foundation models struggle to accurately recognize vehicles and estimate their velocities. Although \methodname{} was not trained on such scenarios, we present a qualitative comparison against the ground truth radar.

\subsection{Additional detection results}
\cref{tab:supp_detection_results} details the detection model's performance on ground truth data compared to synthetic data produced by \methodname{} and the baseline. We evaluate detection on the Car, Truck, and Trailer classes.

\begin{table}[h!]
    \caption{\textbf{Detection metrics comparison.} Evaluation of a trained detector on GT versus generated samples from \methodname{} and Baseline.}
    \label{tab:supp_detection_results}
    \centering
    \resizebox{0.8\columnwidth}{!}{%
    \begin{tabular}{lccccccc}
        \toprule
        Method & mAP $\uparrow$ & mATE $\downarrow$ & mASE $\downarrow$ & mAOE $\downarrow$ & mAVE $\downarrow$ & mAAE $\downarrow$ & NDS $\uparrow$ \\
        \midrule
        GT & 0.38 & 0.54 & 0.18 & 0.11 & 1.88 & 0.24 & 0.48 \\
        RadarGen & 0.11 & 0.91 & 0.20 & 0.21 & 4.04 & 0.19 & 0.30 \\
        Baseline & 0.00 & 1.00 & 1.00 & 1.00 & 1.00 & 1.00 & 0.00 \\
        \bottomrule
    \end{tabular}
    }
\end{table}

In addition, we train VoxelNeXt on both real and synthetic data, with the same configuration as described in \cref{supp:voxelnext-details}, but with an additional augmentation where the input point cloud during training can be either real or synthetic with a 50\% probability. This yields an NDS of 0.48, which is equal to the real only training NDS, when evaluated on real data. This shows our generated data is compatible for training together with real data. We believe augmenting the data by editing available scenes with vehicle replacement can improve results. As this requires a more sophisticated framework for editing multi-view videos, we leave it for future work.

\subsection{nuScenes results}
In addition to MAN TruckScenes\cite{fent2024man} we train and evaluate on the common nuScenes\cite{caesar2020nuscenes} dataset, which includes 5 radar sensors surrounding a driving vehicle. The average point cloud in nuScenes contains only a few hundreds of points, compared to MAN TruckScenes which contain a few thousands.

\myparagraph{Baseline} As for MAN TruckScenes, we use RGB2Point\cite{lee2025rgb2point} as baseline, with the same parameters as described in \cref{supp:baseline-details}. The number of points that can be predicted by the model is constant, therefore we set it to 208, which is approximately the average number of points in the nuScenes dataset.

\myparagraph{Quantitative evaluation} In \cref{tab:pcl_nuscenes_metrics} we compare \methodname{} to the baseline on the nuScenes dataset, with the metrics as described in \cref{sec:supp-metrics-details}. \methodname{} broadly outperforms the baseline.

\input{supp/supp_table_nuscenes_comparison}

\myparagraph{Qualitative evaluation.} In \cref{fig:supp_comparison_nusc}, we visually compare our results with the baseline and ground truth on four examples. \emph{RadarGen's} generated point clouds closely match the ground truth in shape, distribution, and count, demonstrating a significant advantage over the baseline.

\subsection{LiDAR and range image representation}
Many LiDAR generation works \cite{caccia2019deep,sallab2019lidar,xiong2023learning,zyrianov2022learning,ran2024towards,nakashima2024lidar,hu2024rangeldm,wu2024text2lidar} use either range images or rays corresponding to specific range image pixel locations. In this approach, the LiDAR range image resembles a depth map, from which object structure can be inferred visually and the point cloud (PCL) reconstructed easily. Radar can similarly be viewed from the sensor perspective, but issues arise when reconstructing radar range images through a pretrained AE.

As shown in \cref{fig:supp_lidar_radar_comparison}, which compares LiDAR and radar range maps, the radar range map is significantly sparser, making accurate reconstruction harder. \cref{fig:supp_lidar_radar_comparison} also shows a visual comparison of the radar range image before and after passing through SANA's AE, where the degradation is substantial.

In practice, the CD for retrieval of the radar PCL from the original range image is $0.31m$ (before SANA's AE). After reconstruction through the AE the CD is equal to $3.72m$; the MSE between the AE input and output is $0.020$. Both are significantly degraded compared to the errors reported for our BEV map representation in Sec. 5.3 of the manuscript.

Utilizing a LiDAR AE similarly degrades radar's range images. 
LiDAR Diffusion \cite{ran2024towards} is a camera-to-LiDAR method which uses a range image representation with a LiDAR-specific AE and does not output intensity. We tested whether such an AE can be reused for radar by passing the radar range image through its AE: reconstruction degrades from a CD of $0.28m$ (range-image roundtrip alone) to a CD of $4.91m$ after AE encoding/decoding, with substantial artifacts.
Adapting such a method to radar would therefore require designing and training a new AE for sparse, non-uniform radar data and extending the output to RCS and Doppler.

One might consider applying our proposed radar maps mechanism to the range image instead, but this is not possible: pixel values in the range image encode point depth, so applying a Gaussian kernel would discard depth information and corrupt contributions from nearby points.

The reported numbers are averaged over the MAN TruckScenes mini train set. Range images are $96\times1024\text{px}$.

\input{supp/supp_qualitative_open_road}
\input{supp/supp_qualitative_urban}

\input{supp/supp_multiple_seeds}

\input{supp/supp_recovery_methods}

\input{supp/supp_bev_condition}

\input{supp/supp_limitations}

\input{supp/supp_qualitative_nuscenes}

\input{supp/supp_lidar_radar_comparison}

\clearpage

%% file: sections/07_supp_eval.tex
\section{Evaluation Metrics}
\label{sec:supp-metrics-details}

This section provides the detailed formulations for the evaluation metrics introduced in the main paper.

Our evaluation is divided to the \textit{Entire Area} ,which includes the whole point cloud in our range-of-interest ($\pm 50\text{m}$), and the \textit{Foreground}, which is the set  of annotated bounding boxes with (1) camera visibility of more than 60\%, (2) object class in \{Car, Truck, Trailer\}. Comparison is done between evaluated model predictions (synthetic points) and the ground truth points.

\subsection{Chamfer Distance (CD)}
Calculated as the two-way chamfer distance averaged across points and averaged by the two directions:

$$
\text{CD}(P_1, P_2) = \left( d(P_1, P_2) + d(P_2, P_1) \right) / 2~,
$$
where
$$
d(P_1, P_2) = \frac{1}{|P_1|} \sum_{\mathbf{x} \in P_1} \min_{\mathbf{y} \in P_2} \|\mathbf{x} - \mathbf{y}\|_2~.
$$

For CD-Loc we define $P_1$ and $P_2$ as the set of synthetic and ground truth locations  $\{x_i^{syn}, y_i^{syn}\}$, $\{x_i^{gt}, y_i^{gt}\}$ respectively. For CD Full, we define $P_1$ and $P_2$ as the set of normalized locations and attributes $\text{Normalize}(\{x_i,y_i,r_i,d_i\})$, where each value is normalized to the range $[0,1]$ to maintain a similar scale:
$$
u_{\text{normalized}} = \frac{u-u_\text{min}}{u_\text{max}-u_\text{min}}~.
$$

This CD is defined only when there is at least one point in both $P_1$ and $P_2$ and in the \textit{Foreground} it is only calculated on such bounding boxes. While most bounding boxes in our case contain points, this is not always guaranteed. Therefore, we present the Density Similarity metric in \cref{supp:metric_ds} to verify that the similarity of point counts per object matches the ground truth.
We report CD-Loc and CD-Full for the \textit{Entire Area}, averaged across all point cloud pairs, as well as for the \textit{Foreground}, over valid bounding boxes.

\subsection{Point Cloud IoU@1m}
We adopt the definition from \cite{saunders2024baseboostdepth}, which computes the Intersection over Union (IoU) between two point clouds, $P_1$ and $P_2$, which are sets of locations $\{x_i^{syn}, y_i^{syn}\}$, $\{x_i^{gt}, y_i^{gt}\}$ respectively, using a matching distance threshold $\delta$.
$$
IoU = \frac{P\cdot R}{P+R-P\cdot R}~,
$$
where precision and recall
\begin{equation*}
\begin{array}{lcl}
P & = & \frac{1}{|P_1|} \sum_{p_1 \in P_1} \mathbb{I} 
\left[ \min_{p_2 \in P_2} \| p_2 - p_1 \|_2 < \delta \right], \\
R & = & \frac{1}{|P_2|} \sum_{p_2 \in P_2} \mathbb{I} \left[ \min_{p_1 \in P_1} \| p_1 - p_2 \|_2 < \delta \right].
\end{array}
\end{equation*}
In our setting, we set $\delta=1m$. We report IoU@1m for the \textit{Entire Area}, averaged across all point cloud pairs.

\subsection{Density Similarity}
\label{supp:metric_ds}
For the point cloud inside each bounding box we measure the difference in number of points compared to the ground truth points in that box.
Define $N,M$ as the number of points in point clouds $P_1, P_2$ respectively. Then the density similarity between $P_1$ and $P_2$ is defined as:

$$
DS(P_1, P_2) = 
\begin{cases} 
  1 & \text{if } N = 0 \text{ and } M = 0 \\
  \frac{\min(N, M)}{\max(N, M)} & \text{otherwise}
\end{cases}~.
$$
We report Density Similarity for the \textit{Foreground}, averaged across all bounding boxes.

\subsection{Bounding Box Hit Rate}
We consider the set of bounding boxes that contain at least one ground-truth point. The Hit Rate is the fraction of these bounding boxes that also contain at least one synthetic point.

$$
\text{Hit Rate} = \frac{\text{Bounding boxes with synthetic points}}{\text{Bounding boxes with ground truth points}}~.
$$
We report Hit Rate for the \textit{Foreground}, averaged across all bounding boxes.

\subsection{Distance-Attribute (DA)}
We propose the Distance-Attribute (DA) metric to jointly evaluate the accuracy of point locations and attributes. We define a "hit" only if a synthetic point falls within specific difference thresholds for spatial distance, RCS, and Doppler relative to a ground truth point. Unlike high-dimensional Chamfer Distance (CD-Full), DA enforces strict spatial locality, ensuring that spatially distant points with similar attributes are not matched. To avoid ambiguity arising from point ordering in greedy approaches, we solve a global assignment problem to identify the set of pairs satisfying all conditions. We define:
\begin{equation*}
\begin{array}{lcl}
TP & = & \text{\# of matched pairs}, \\
FN & = & \text{\# of unmatched GT points}, \\
FP & = & \text{\# of unmatched synthetic points},
\end{array}
\end{equation*}
from which Precision, Recall, and F1 are derived. In our setting, we select the thresholds: $\delta_{loc}=1\text{m}$, $\delta_{RCS}=8\text{dBsm}$, and $\delta_{Doppler}=2.5\text{m/s}$.
We report DA Recall, Precision, and F1 for the \textit{Entire Area}, averaged across all point cloud pairs.

\subsection{Maximum Mean Discrepancy (MMD)}
\label{sec:supp-metrics-mmd}

To measure the distributional similarity between the synthetic point cloud $P_1$ and the ground truth $P_2$, we utilize Maximum Mean Discrepancy (MMD) \cite{gretton2012kernel}. We compute MMD independently for each point cloud attribute: location  $P_i^\text{Loc}$, RCS $P_i^\text{RCS}$ and Doppler $P_i^\text{Doppler}$. We employ a multi-scale Radial Basis Function (RBF) kernel defined as a sum of $K=5$ Gaussians:
$$k(u, v) = \sum_{l=1}^{K} \exp\left(-\frac{\|u - v\|^2}{h_l}\right).$$
The kernel bandwidths are set as $h_l = h_{\text{base}} \cdot 2^{l-3}$, where the base bandwidth $h_{\text{base}}$ is calculated as the mean squared Euclidean distance between all distinct pairs in the joint set $P_1^\text{attr} \cup P_2^\text{attr}$ for a given attribute. The final MMD metric is computed as the biased estimator over the resulting kernel matrix.

\myparagraph{\textbf{Entire Area.}} For each pair of synthetic and ground truth point clouds, we report MMD separately on location, RCS, and Doppler.

\myparagraph{\textbf{Foreground.}}
For each class \{Car, Truck, Trailer\}, we aggregate points from annotated bounding boxes after transforming them into a canonical coordinate system via centering and rotation alignment. We report the MMD between the aggregated synthetic and ground truth sets for each class, calculated separately for location, RCS, and Doppler.

%% file: supp/supp_table_nuscenes_comparison.tex
\begin{table*}[t]
\centering
\caption{
    \textbf{Quantitative evaluation.}
    \methodname{} broadly outperforms the baseline on geometric fidelity (CD, IoU, Density Similarity, Hit Rate), radar attribute fidelity (DA Recall, Precision, F1), and distribution similarity (MMD) on the \textbf{nuScenes}\cite{caesar2020nuscenes} dataset. Shown is the mean $\pm$ s.d., computed over all timestamps for the Entire Area and over all foreground objects for the Foreground.
}
\vspace{-0.5em}
\label{tab:pcl_nuscenes_metrics}
\resizebox{\textwidth}{!}{%
\begin{tabular}{lccccccccc}
\toprule
\multirow{2}{*}{\textbf{Method}} & \multicolumn{9}{c}{\textbf{Entire Area}} \\
\cmidrule(lr){2-10}
& \multirow{1}{*}{\textbf{CD Loc. ($\downarrow$)}} & \multirow{1}{*}{\textbf{CD Full ($\downarrow$)}} & \multirow{1}{*}{\textbf{IoU@1m ($\uparrow$)}} & \multirow{1}{*}{\textbf{DA Recall ($\uparrow$)}} & \multirow{1}{*}{\textbf{DA Prec. ($\uparrow$)}} & \multirow{1}{*}{\textbf{DA F1 ($\uparrow$)}} & \multirow{1}{*}{\textbf{MMD Loc. ($\downarrow$)}} & \multirow{1}{*}{\textbf{MMD RCS ($\downarrow$)}} & \multirow{1}{*}{\textbf{MMD Doppler ($\downarrow$)}} \\
\midrule
Baseline & $3.85 \pm 0.85$          & $\mathbf{0.062 \pm 0.010}$ & $0.04 \pm 0.04$          & $0.04 \pm 0.03$            & $0.02 \pm 0.02$          & $0.03 \pm 0.02$          & $0.878 \pm 0.238$          & $0.70 \pm 0.30$ & $1.34 \pm 1.10$ \\
RadarGen & $\mathbf{3.50 \pm 0.85}$ & $0.063 \pm 0.011$          & $\mathbf{0.13 \pm 0.06}$ & $\mathbf{0.14 \pm 0.08}$   & $\mathbf{0.05 \pm 0.04}$ & $\mathbf{0.08 \pm 0.05}$ & $\mathbf{0.093 \pm 0.083}$ & $\mathbf{0.16 \pm 0.18}$ & $\mathbf{0.55 \pm 0.68}$ \\
\bottomrule
\end{tabular}
}

\resizebox{\textwidth}{!}{%
\begin{tabular}{lccccccccccccc} 
\multirow{3}{*}{\textbf{}} & \multicolumn{13}{c}{\textbf{Foreground}} \\ 
\cmidrule(lr){2-14} 
& \multirow{2}{*}{\textbf{CD Loc. ($\downarrow$)}} & \multirow{2}{*}{\textbf{CD Full ($\downarrow$)}} & \multirow{2}{*}{\textbf{Density Sim. ($\uparrow$)}} & \multirow{2}{*}{\textbf{Hit Rate ($\uparrow$)}} & \multicolumn{3}{c}{\textbf{MMD Car ($\downarrow$)}} & \multicolumn{3}{c}{\textbf{MMD Truck ($\downarrow$)}} & \multicolumn{3}{c}{\textbf{MMD Trailer ($\downarrow$)}} \\ 
\cmidrule(lr){6-8} \cmidrule(lr){9-11} \cmidrule(lr){12-14} 
& & & & & \textbf{Loc.} & \textbf{RCS} & \textbf{Doppler} & \textbf{Loc.} & \textbf{RCS} & \textbf{Doppler} & \textbf{Loc.} & \textbf{RCS} & \textbf{Doppler} \\ 
\midrule
Baseline & $1.83 \pm 1.24$          & $0.063 \pm 0.039$          & $0.64 \pm 0.47$          & $0.18$          & $0.182$           & $1.108$          & $0.670$          & $0.190$          & $1.132$          & $0.805$          & $\mathbf{0.013}$  & $0.227$          & $0.108$ \\
RadarGen & $\mathbf{1.30 \pm 1.04}$ & $\mathbf{0.059 \pm 0.040}$ & $\mathbf{0.65 \pm 0.44}$ & $\mathbf{0.49}$ & $\mathbf{0.110}$  & $\mathbf{0.004}$ & $\mathbf{0.043}$ & $\mathbf{0.014}$ & $\mathbf{0.014}$ & $\mathbf{0.116}$ & $0.046$           & $\mathbf{0.056}$ & $\mathbf{0.032}$ \\
\bottomrule
\end{tabular}
}
\vspace{-0.5em}
\end{table*}

%% file: supp/supp_qualitative_open_road.tex
\begin{figure*}[!p]
  \centering
  \includegraphics[width=1\linewidth]{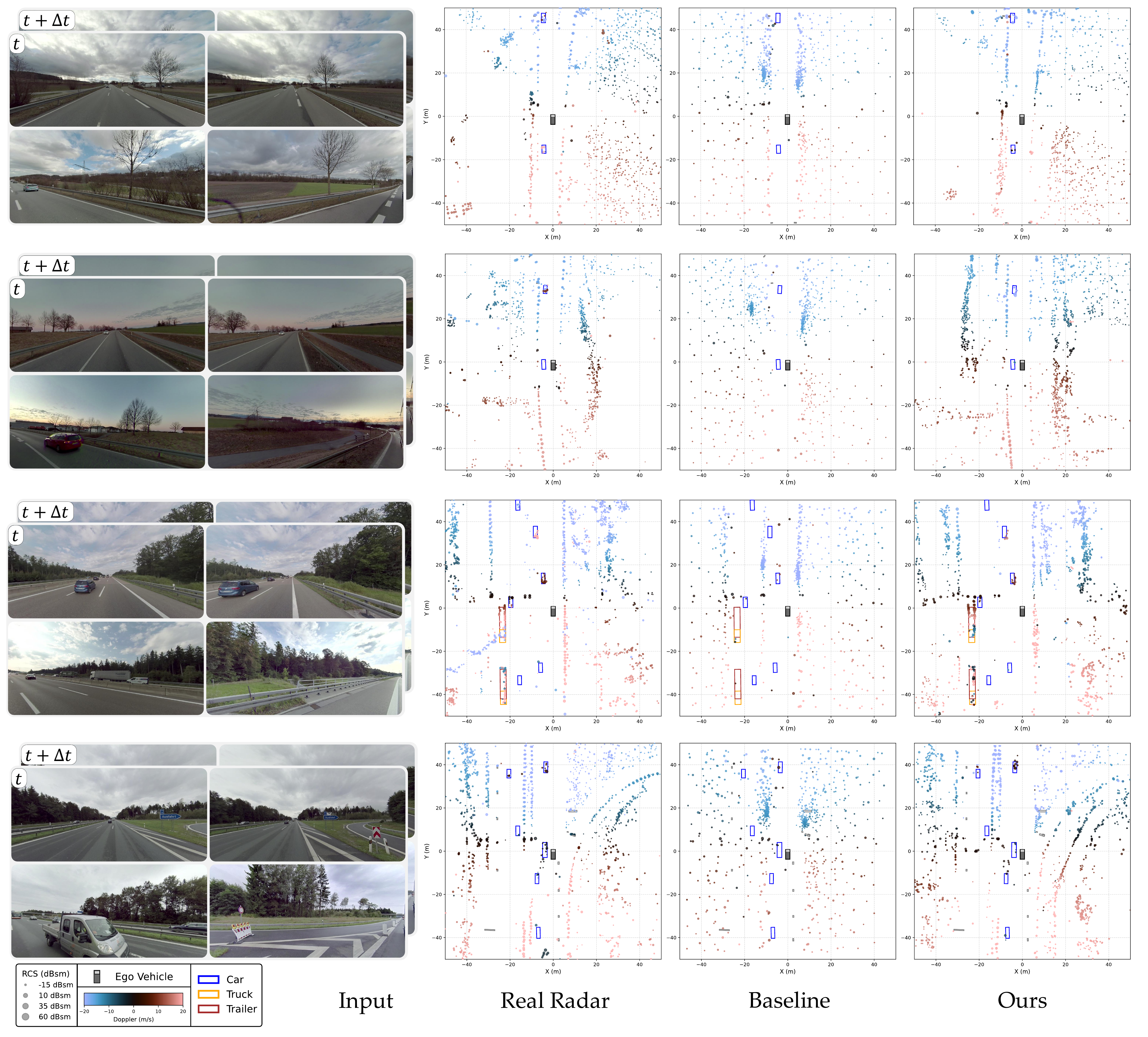}
  \vspace{-1.5em}
  \caption{\textbf{Additional qualitative results.} Additional demonstration of \methodname{} compared to the baseline and ground truth in highway and rural scenarios. Our model generates point clouds with higher geometric and attribute fidelity to the ground truth compared to the baseline. \methodname{} uses inputs $t$ and $t+\Delta t$, while the baseline uses only $t$. Ground truth bounding boxes are highlighted in color.}
  \label{fig:supp_comparison_open_road}
\end{figure*}

%% file: supp/supp_qualitative_urban.tex
\begin{figure*}[!p]
  \centering
  \includegraphics[width=1\linewidth]{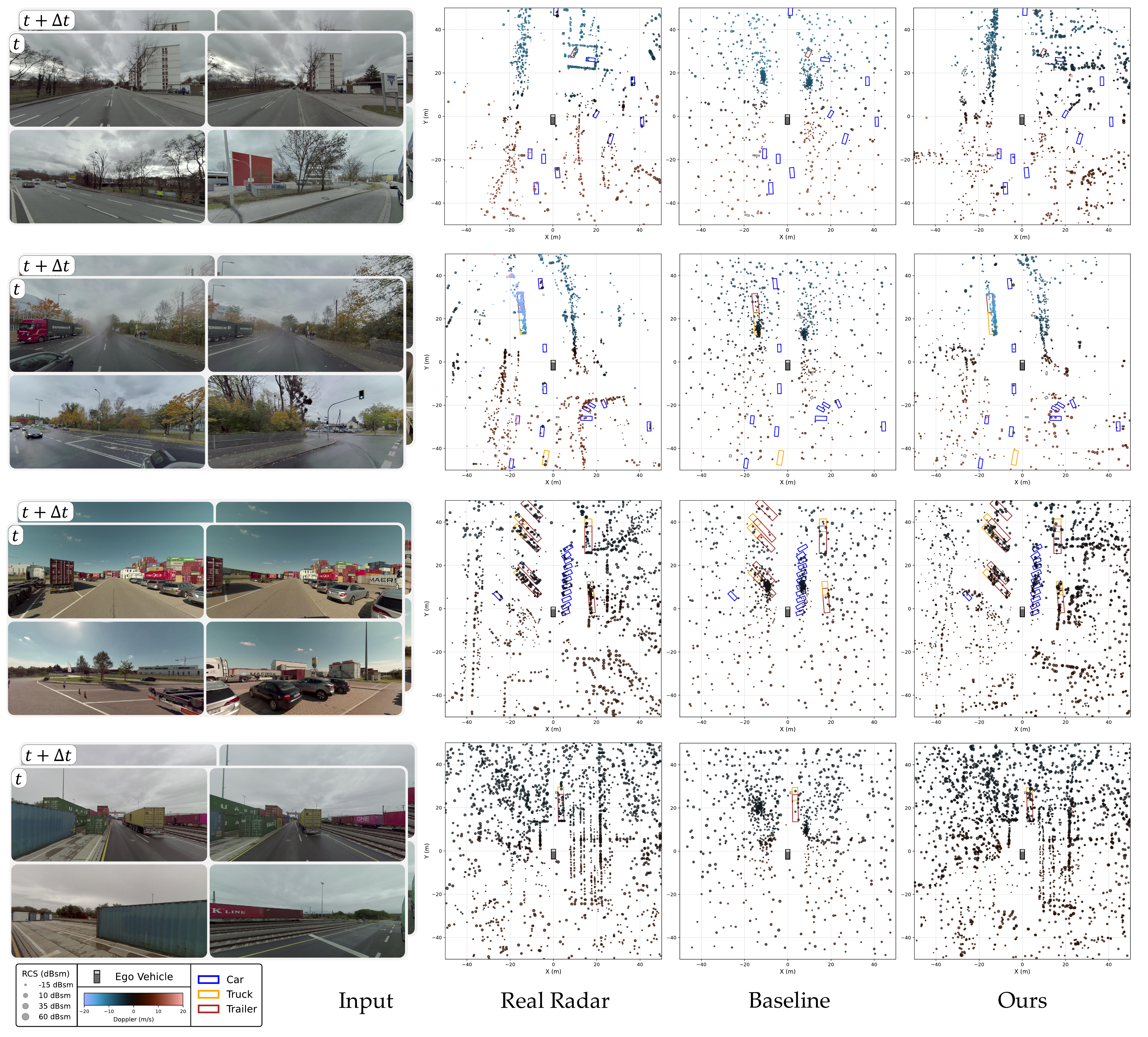}
  \vspace{-1.5em}
  \caption{\textbf{Additional qualitative results.} Additional demonstration of \methodname{} compared to the baseline and ground truth in urban environments. Our model generates point clouds with higher geometric and attribute fidelity to the ground truth compared to the baseline. \methodname{} uses inputs $t$ and $t+\Delta t$, while the baseline uses only $t$. Ground truth bounding boxes are highlighted in color.}
  \label{fig:supp_comparison_urban}
\end{figure*}

%% file: supp/supp_multiple_seeds.tex
\begin{figure*}[!p]
  \centering
  \includegraphics[width=1\linewidth]{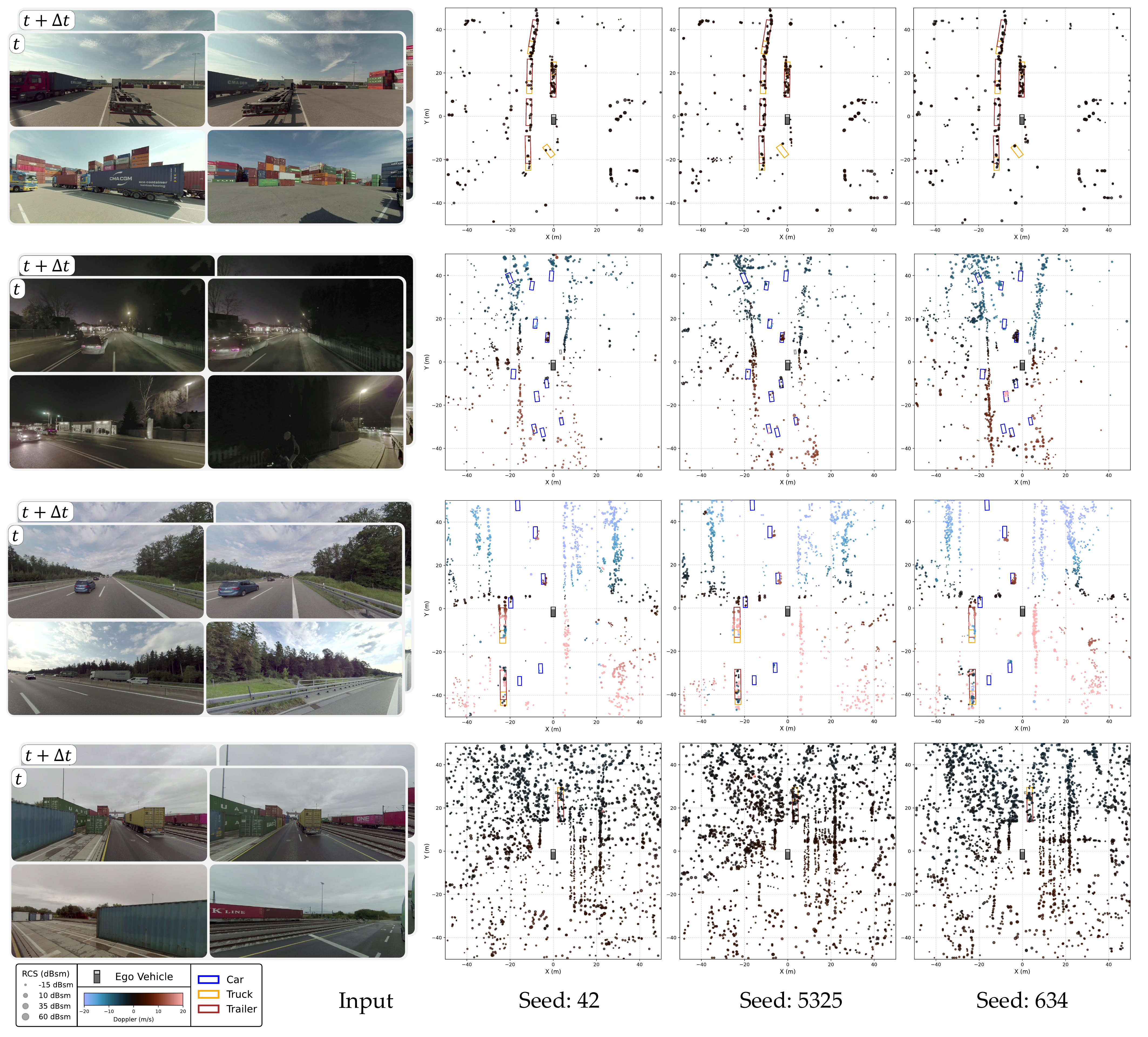}
  \vspace{-1.5em}
  \caption{\textbf{Additional seeds.} Our model can generate multiple sets of point clouds for a single scene by replacing the diffusion process seed. \methodname{} uses inputs $t$ and $t+\Delta t$. Ground truth bounding boxes are highlighted in color.}
  \label{fig:supp_multiple_seeds}
\end{figure*}

%% file: supp/supp_recovery_methods.tex
\begin{figure*}[!p]
  \centering
  \includegraphics[width=1\linewidth]{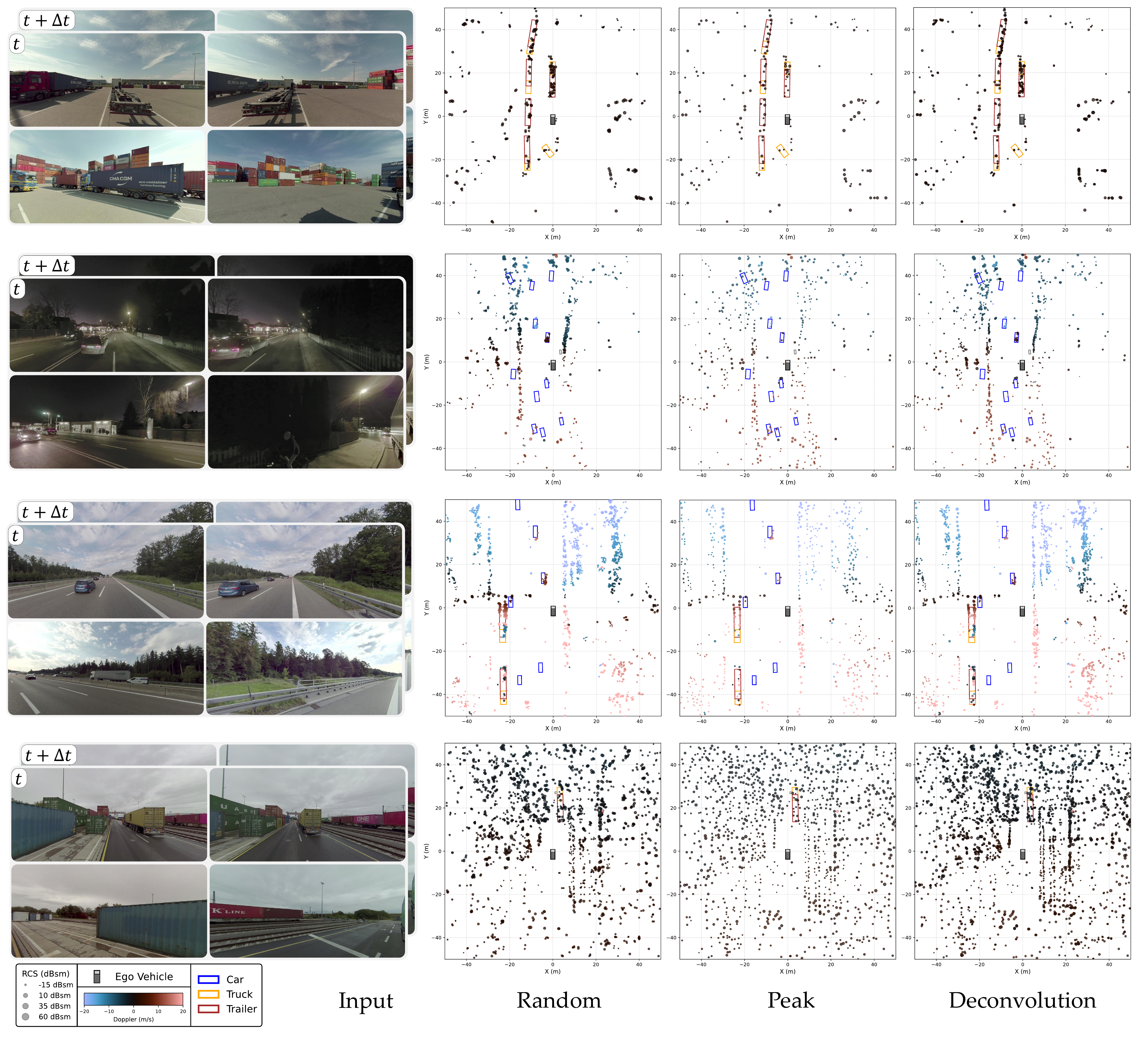}
  \vspace{-1.5em}
  \caption{\textbf{Recovery methods.} Comparison of Random, Peak, and Deconvolution sparse point cloud recovery methods. Random sampling exhibits inconsistent density characterized by clustering and empty regions. Peak recovery fills the space uniformly but suffers from low density. Our Deconvolution method achieves coverage while maintaining density where necessary. \methodname{} uses inputs $t$ and $t+\Delta t$. Ground truth bounding boxes are highlighted in color.}
  \label{fig:supp_recovery}
\end{figure*}

%% file: supp/supp_bev_condition.tex
\begin{figure*}[!p]
  \centering
  \includegraphics[width=1\linewidth]{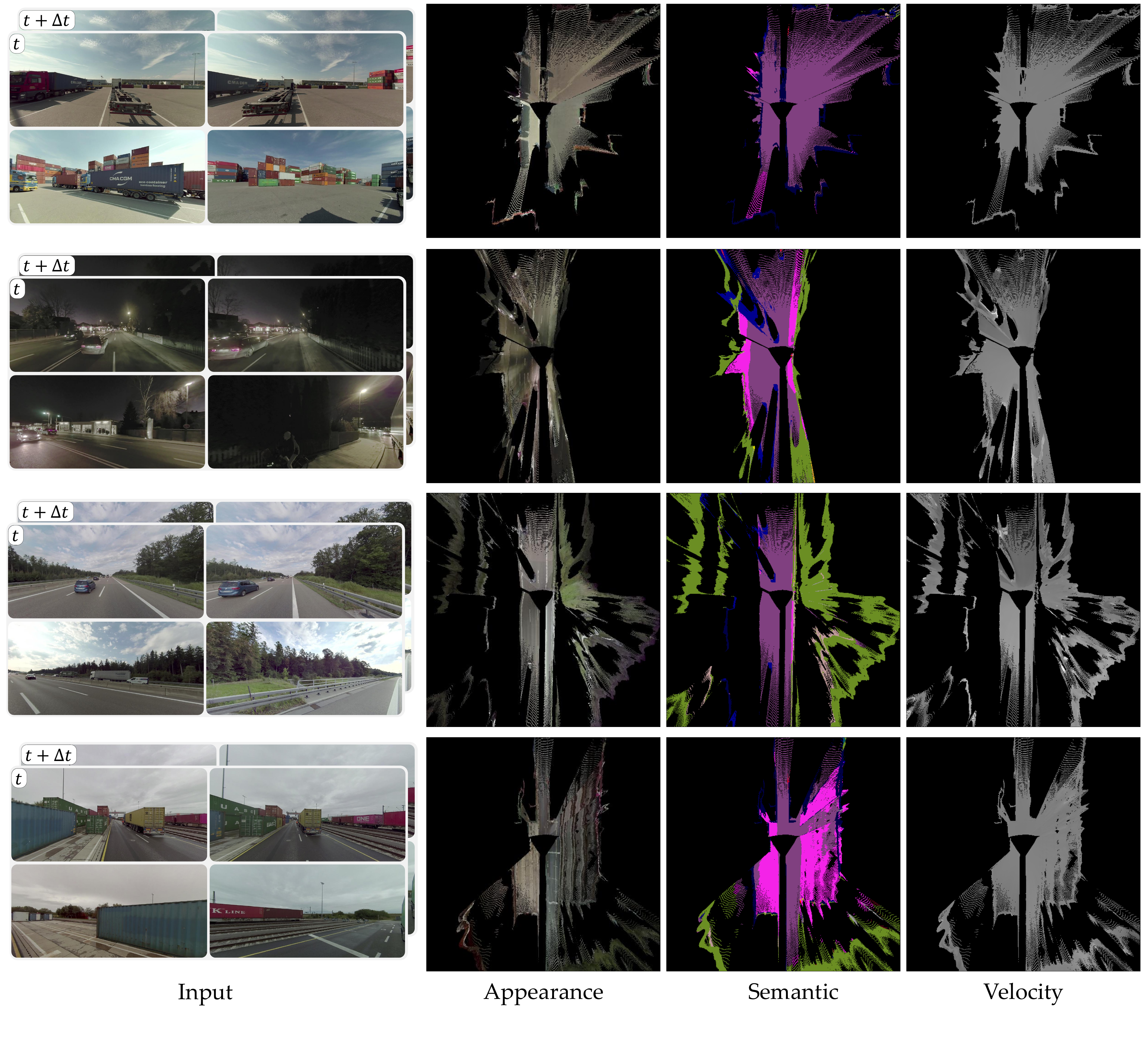}
  \vspace{-1.5em}
  \caption{\textbf{BEV conditioning maps.} Visualization of the BEV appearance, semantic, and relative radial velocity maps produced from inputs at times $t$ and $t+\Delta t$. The appearance map retains the camera image colors. Semantic classes are color-coded as: \cbox{road}~Road, \cbox{sidewalk}~Sidewalk, \cbox{building}~Building, \cbox{vegetation}~Vegetation, \cbox{car}~Car, and \cbox{person}~Person. For the velocity map, lighter colors indicate positive velocity while darker colors indicate negative velocity.}
  \label{fig:supp_bev_condition}
\end{figure*}

%% file: supp/supp_limitations.tex
\begin{figure*}[!p]
  \centering
  \includegraphics[width=1\linewidth]{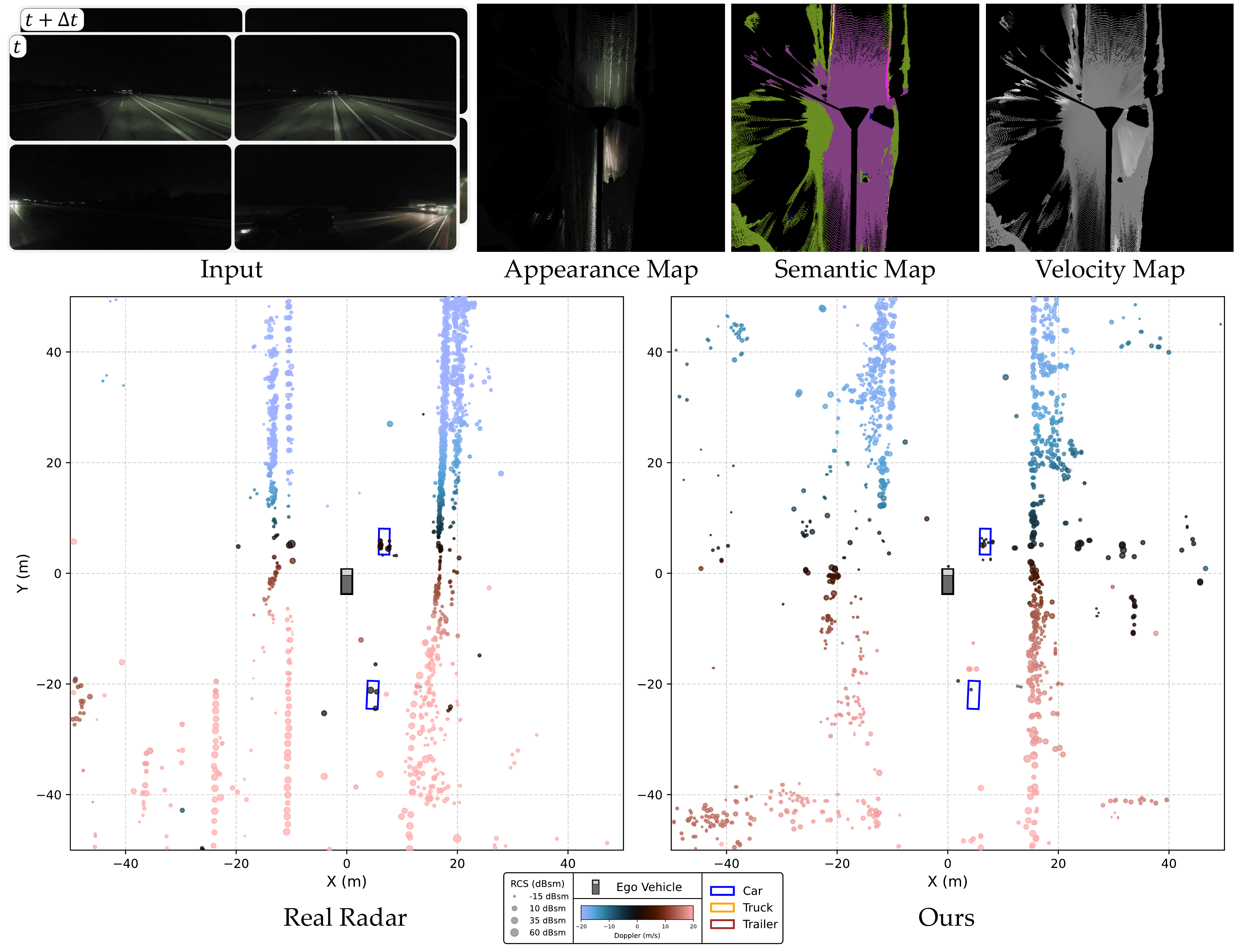}
  \vspace{-1.5em}
  \caption{\textbf{Qualitative analysis of limitations.} Visual comparison of \methodname{} against the ground truth radar in a low-light night scene. In this setting, the underlying foundation models struggle to accurately recognize vehicles and estimate velocities. \methodname{} was not trained on such scenarios. Ground truth bounding boxes are highlighted in color.}
  \label{fig:supp_limitation}
\end{figure*}

%% file: supp/supp_qualitative_nuscenes.tex
\begin{figure*}[!p]
  \centering
  \includegraphics[width=1\linewidth]{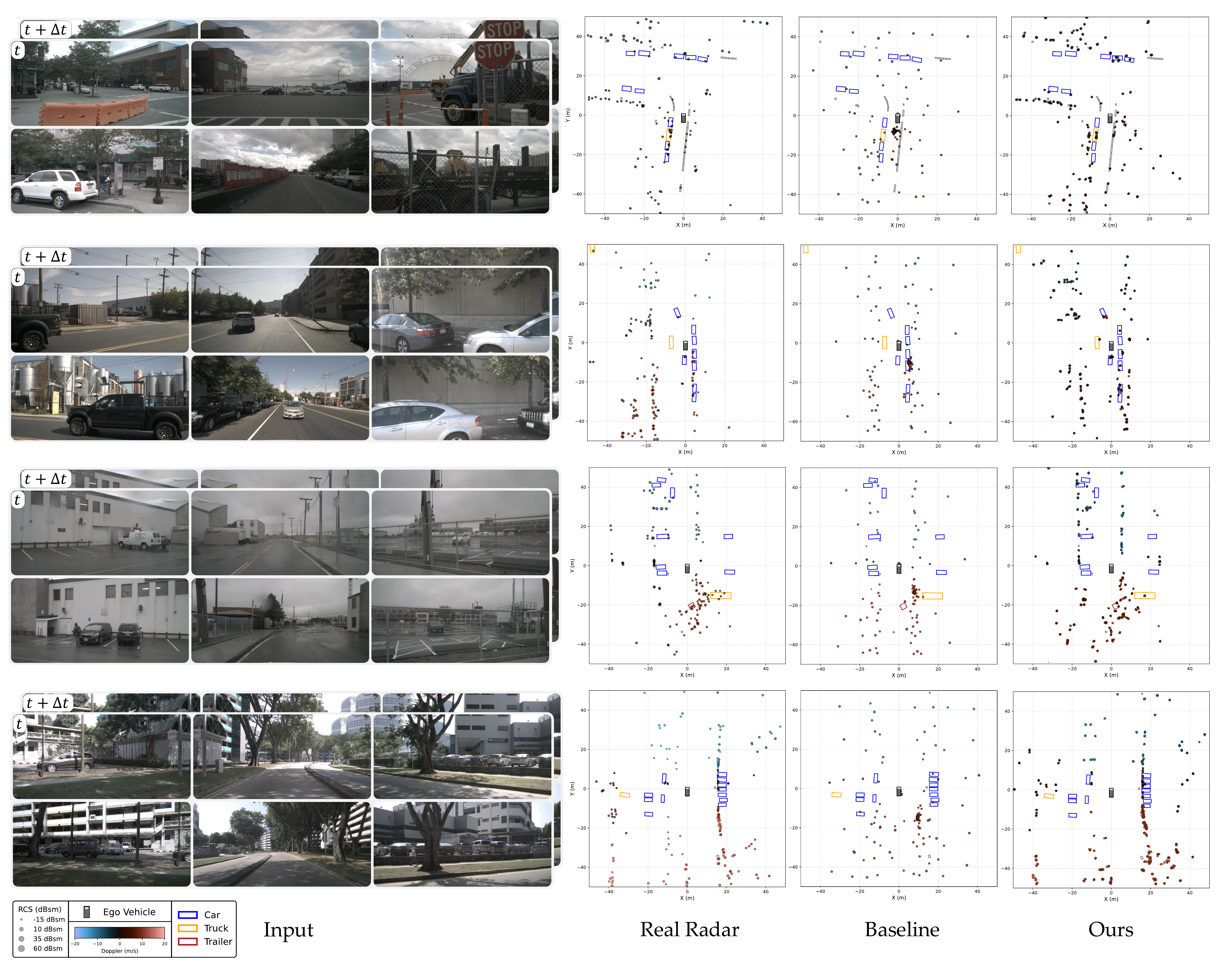}
  \vspace{-1.5em}
  \caption{\textbf{NuScenes qualitative results.}  Our model generates point clouds with higher geometric and attribute fidelity to the ground truth compared to the baseline. \methodname{} uses inputs $t$ and $t+\Delta t$, while the baseline uses only $t$. Ground truth bounding boxes are highlighted in color.}
  \label{fig:supp_comparison_nusc}
\end{figure*}

%% file: supp/supp_lidar_radar_comparison.tex
\begin{figure*}[!p]
  \centering
  \includegraphics[width=1\linewidth]{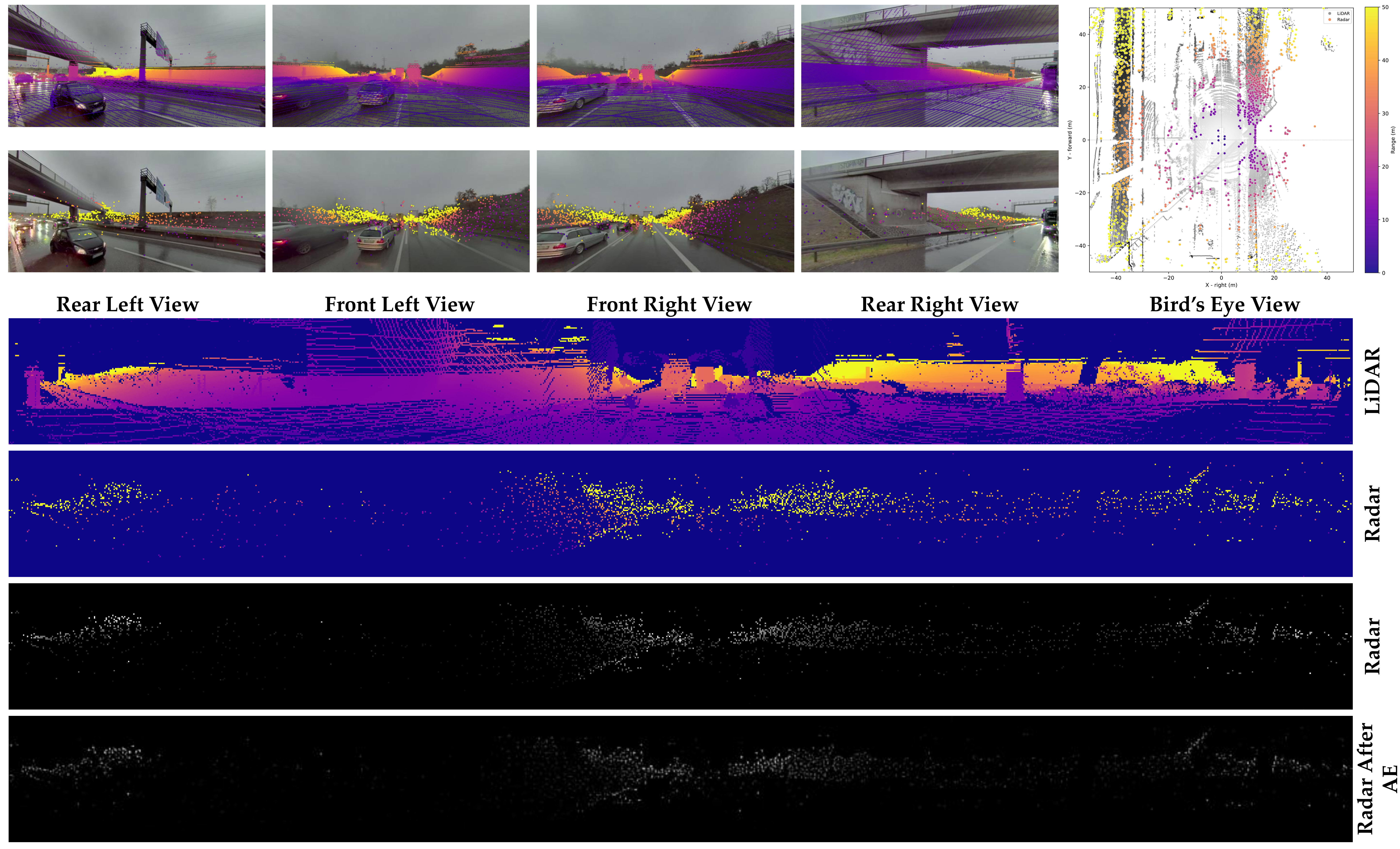}
  \vspace{-1.5em}
  \caption{\textbf{Range images.} Top: comparison of LiDAR and radar point clouds in camera and bird’s eye views. Middle: LiDAR and radar range images are shown in color for convenience, with the same color scheme as presented at the top. The radar point cloud is much sparser than the LiDAR, therefore the image is not as dense. Bottom: Radar range image in grayscale as used in practice for the AE. Beneath is the reconstructed radar range image after being encoded and decoded through SANA's AE. In this specific example the CD between the original radar PCL and radar PCL retrieved from the range image is $0.21m$, after the AE reconstruction the CD is $2.97m$, significantly degraded.}
  \label{fig:supp_lidar_radar_comparison}
\end{figure*}

%% file: main.bbl
\begin{thebibliography}{10}
\providecommand{\url}[1]{\texttt{#1}}
\providecommand{\urlprefix}{URL }
\providecommand{\doi}[1]{https://doi.org/#1}

\bibitem{achlioptas2018learning}
Achlioptas, P., Diamanti, O., Mitliagkas, I., Guibas, L.: Learning representations and generative models for 3d point clouds. In: ICML (2018)

\bibitem{10447724}
Alkanat, T., Pandharipande, A.: Automotive radar point cloud parametric density estimation using camera images. In: ICASSP (2024)

\bibitem{arnold2022maxray}
Arnold, M., Bauhofer, M., Mandelli, S., Henninger, M., Schaich, F., Wild, T., ten Brink, S.: Maxray: A raytracing-based integrated sensing and communication framework. In: IEEE JC\&S (2022)

\bibitem{beck2009fast}
Beck, A., Teboulle, M.: A fast iterative shrinkage-thresholding algorithm for linear inverse problems. SIAM journal on imaging sciences  (2009)

\bibitem{bialer2024radsimreal}
Bialer, O., Haitman, Y.: Radsimreal: Bridging the gap between synthetic and real data in radar object detection with simulation. In: CVPR (2024)

\bibitem{borts2024radar}
Borts, D., Liang, E., Broedermann, T., Ramazzina, A., Walz, S., Palladin, E., Sun, J., Brueggemann, D., Sakaridis, C., Van~Gool, L., et~al.: Radar fields: Frequency-space neural scene representations for fmcw radar. In: ACM SIGGRAPH (2024)

\bibitem{caccia2019deep}
Caccia, L., Van~Hoof, H., Courville, A., Pineau, J.: Deep generative modeling of lidar data. In: IROS (2019)

\bibitem{caesar2020nuscenes}
Caesar, H., Bankiti, V., Lang, A.H., Vora, S., Liong, V.E., Xu, Q., Krishnan, A., Pan, Y., Baldan, G., Beijbom, O.: nuscenes: A multimodal dataset for autonomous driving. In: CVPR (2020)

\bibitem{cai2020learning}
Cai, R., Yang, G., Averbuch-Elor, H., Hao, Z., Belongie, S., Snavely, N., Hariharan, B.: Learning gradient fields for shape generation. In: ECCV (2020)

\bibitem{capsoni1998physically}
Capsoni, C., D’Amico, M.: A physically based radar simulator. Journal of Atmospheric and Oceanic Technology  (1998)

\bibitem{chen2023rf}
Chen, X., Zhang, X.: Rf genesis: Zero-shot generalization of mmwave sensing through simulation-based data synthesis and generative diffusion models. In: Proceedings of the 21st ACM Conference on Embedded Networked Sensor Systems (2023)

\bibitem{chen2023voxelnext}
Chen, Y., Liu, J., Zhang, X., Qi, X., Jia, J.: Voxelnext: Fully sparse voxelnet for 3d object detection and tracking. In: CVPR (2023)

\bibitem{cheng2021mask2former}
Cheng, B., Misra, I., Schwing, A.G., Kirillov, A., Girdhar, R.: Masked-attention mask transformer for universal image segmentation. In: CVPR (2022)

\bibitem{cordts2016cityscapes}
Cordts, M., Omran, M., Ramos, S., Rehfeld, T., Enzweiler, M., Benenson, R., Franke, U., Roth, S., Schiele, B.: The cityscapes dataset for semantic urban scene understanding. In: CVPR (2016)

\bibitem{daubechies2010iteratively}
Daubechies, I., DeVore, R., Fornasier, M., G{\"u}nt{\"u}rk, C.S.: Iteratively reweighted least squares minimization for sparse recovery. Communications on Pure and Applied Mathematics: A Journal Issued by the Courant Institute of Mathematical Sciences  (2010)

\bibitem{de2020generating}
De~Oliveira, M.L.L., Bekooij, M.J.: Generating synthetic short-range fmcw range-doppler maps using generative adversarial networks and deep convolutional autoencoders. In: IEEE Radar Conference (2020)

\bibitem{deng2023midas}
Deng, K., Zhao, D., Han, Q., Zhang, Z., Wang, S., Zhou, A., Ma, H.: Midas: Generating mmwave radar data from videos for training pervasive and privacy-preserving human sensing tasks. Proceedings of the ACM on IMWUT  (2023)

\bibitem{ding2023hidden}
Ding, F., Palffy, A., Gavrila, D.M., Lu, C.X.: Hidden gems: 4d radar scene flow learning using cross-modal supervision. In: CVPR (2023)

\bibitem{ding2022self}
Ding, F., Pan, Z., Deng, Y., Deng, J., Lu, C.X.: Self-supervised scene flow estimation with 4-d automotive radar. IEEE RA-L  (2022)

\bibitem{ding2024radarocc}
Ding, F., Wen, X., Zhu, Y., Li, Y., Lu, C.X.: Radarocc: Robust 3d occupancy prediction with 4d imaging radar. NeurIPS  (2024)

\bibitem{dosovitskiy2017carla}
Dosovitskiy, A., Ros, G., Codevilla, F., Lopez, A., Koltun, V.: Carla: An open urban driving simulator. In: Conference on robot learning (2017)

\bibitem{dudek2011millimeter}
Dudek, M., Kissinger, D., Weigel, R., Fischer, G.: A millimeter-wave fmcw radar system simulator for automotive applications including nonlinear component models. In: EuRAD (2011)

\bibitem{fent2024man}
Fent, F., Kuttenreich, F., Ruch, F., Rizwin, F., Juergens, S., Lechermann, L., Nissler, C., Perl, A., Voll, U., Yan, M., et~al.: Man truckscenes: A multimodal dataset for autonomous trucking in diverse conditions. NeurIPS  (2024)

\bibitem{fidelis2023generation}
Fidelis, E.C., Reway, F., Ribeiro, H., Campos, P.L., Huber, W., Icking, C., Faria, L.A., Sch{\"o}n, T.: Generation of realistic synthetic raw radar data for automated driving applications using generative adversarial networks. arXiv preprint arXiv:2308.02632  (2023)

\bibitem{giannakis2015realistic}
Giannakis, I., Giannopoulos, A., Warren, C.: A realistic fdtd numerical modeling framework of ground penetrating radar for landmine detection. IEEE journal of selected topics in applied earth observations and remote sensing  (2015)

\bibitem{goodfellow2020generative}
Goodfellow, I., Pouget-Abadie, J., Mirza, M., Xu, B., Warde-Farley, D., Ozair, S., Courville, A., Bengio, Y.: Generative adversarial networks. Communications of the ACM  (2020)

\bibitem{gretton2012kernel}
Gretton, A., Borgwardt, K.M., Rasch, M.J., Sch{\"o}lkopf, B., Smola, A.: A kernel two-sample test. JMLR  (2012)

\bibitem{gubelli2013ray}
Gubelli, D., Krasnov, O.A., Yarovyi, O.: Ray-tracing simulator for radar signals propagation in radar networks. In: EuRAD (2013)

\bibitem{he2022channel}
He, D., Guan, K., Ai, B., Zhong, Z., Kim, J., Chung, H., Hrovat, A.: Channel measurement and ray-tracing simulation for 77 ghz automotive radar. IEEE TITS  (2022)

\bibitem{he2025unirelight}
He, K., Liang, R., Munkberg, J., Hasselgren, J., Vijaykumar, N., Keller, A., Fidler, S., Gilitschenski, I., Gojcic, Z., Wang, Z.: Unirelight: Learning joint decomposition and synthesis for video relighting. arXiv preprint arXiv:2506.15673  (2025)

\bibitem{hirsenkorn2017ray}
Hirsenkorn, N., Subkowski, P., Hanke, T., Schaermann, A., Rauch, A., Rasshofer, R., Biebl, E.: A ray launching approach for modeling an fmcw radar system. In: IRS (2017)

\bibitem{ho2020denoising}
Ho, J., Jain, A., Abbeel, P.: Denoising diffusion probabilistic models. arXiv preprint arXiv:2006.11239  (2020)

\bibitem{holder2019fourier}
Holder, M., Linnhoff, C., Rosenberger, P., Winner, H.: The fourier tracing approach for modeling automotive radar sensors. In: IRS (2019)

\bibitem{hu2024rangeldm}
Hu, Q., Zhang, Z., Hu, W.: Rangeldm: Fast realistic lidar point cloud generation. In: ECCV (2024)

\bibitem{huang2024dart}
Huang, T., Miller, J., Prabhakara, A., Jin, T., Laroia, T., Kolter, Z., Rowe, A.: Dart: Implicit doppler tomography for radar novel view synthesis. In: CVPR (2024)

\bibitem{ke2024repurposing}
Ke, B., Obukhov, A., Huang, S., Metzger, N., Daudt, R.C., Schindler, K.: Repurposing diffusion-based image generators for monocular depth estimation. In: CVPR (2024)

\bibitem{kerbl20233d}
Kerbl, B., Kopanas, G., Leimk{\"u}hler, T., Drettakis, G.: 3d gaussian splatting for real-time radiance field rendering. ACM Trans. Graph.  (2023)

\bibitem{kingma2019introduction}
Kingma, D.P., Welling, M., et~al.: An introduction to variational autoencoders. Foundations and Trends{\textregistered} in Machine Learning  (2019)

\bibitem{kung2025radarsplat}
Kung, P.C., Harisha, S., Vasudevan, R., Eid, A., Skinner, K.A.: Radarsplat: Radar gaussian splatting for high-fidelity data synthesis and 3d reconstruction of autonomous driving scenes. arXiv preprint arXiv:2506.01379  (2025)

\bibitem{lee2025rgb2point}
Lee, J.J., Benes, B.: Rgb2point: 3d point cloud generation from single rgb images. In: WACV (2025)

\bibitem{lei2024sar}
Lei, Z., Xu, F., Wei, J., Cai, F., Wang, F., Jin, Y.Q.: Sar-nerf: Neural radiance fields for synthetic aperture radar multi-view representation. IEEE TGRS  (2024)

\bibitem{li2025sar}
Li, A., Lei, Z., Wei, J., Xu, F.: Sar-gs: 3d gaussian splatting for synthetic aperture radar target reconstruction. arXiv preprint arXiv:2506.21633  (2025)

\bibitem{WeatherWeaver}
Lin, C.H., Wang, Z., Liang, R., Zhang, Y., Fidler, S., Wang, S., Gojcic, Z.: Controllable weather synthesis and removal with video diffusion models. IEEE/CVF International Conference on Computer Vision (ICCV)  (2025)

\bibitem{liu2019simulation}
Liu, H., Xing, B., Wang, H., Cui, J., Spencer, B.F.: Simulation of ground penetrating radar on dispersive media by a finite element time domain algorithm. Journal of applied geophysics  (2019)

\bibitem{luo2021diffusion}
Luo, S., Hu, W.: Diffusion probabilistic models for 3d point cloud generation. In: CVPR (2021)

\bibitem{melas2023pc2}
Melas-Kyriazi, L., Rupprecht, C., Vedaldi, A.: Pc2: Projection-conditioned point cloud diffusion for single-image 3d reconstruction. In: CVPR (2023)

\bibitem{mildenhall2021nerf}
Mildenhall, B., Srinivasan, P.P., Tancik, M., Barron, J.T., Ramamoorthi, R., Ng, R.: Nerf: Representing scenes as neural radiance fields for view synthesis. Communications of the ACM  (2021)

\bibitem{mo2019structurenet}
Mo, K., Guerrero, P., Yi, L., Su, H., Wonka, P., Mitra, N.J., Guibas, L.J.: Structurenet: hierarchical graph networks for 3d shape generation. ACM TOG  (2019)

\bibitem{nakashima2024lidar}
Nakashima, K., Kurazume, R.: Lidar data synthesis with denoising diffusion probabilistic models. In: ICRA (2024)

\bibitem{nichol2022point}
Nichol, A., Jun, H., Dhariwal, P., Mishkin, P., Chen, M.: Point-e: A system for generating 3d point clouds from complex prompts. arXiv preprint arXiv:2212.08751  (2022)

\bibitem{nvidia2025cosmosworldfoundationmodel}
NVIDIA, Agarwal, N., Ali, A., Bala, M., Balaji, Y., Barker, E., Cai, T., Chattopadhyay, P., Chen, Y., Cui, Y., Ding, Y., Dworakowski, D., Fan, J., Fenzi, M., Ferroni, F., Fidler, S., Fox, D., Ge, S., Ge, Y., Gu, J., Gururani, S., He, E., Huang, J., Huffman, J., Jannaty, P., Jin, J., Kim, S.W., Klár, G., Lam, G., Lan, S., Leal-Taixe, L., Li, A., Li, Z., Lin, C.H., Lin, T.Y., Ling, H., Liu, M.Y., Liu, X., Luo, A., Ma, Q., Mao, H., Mo, K., Mousavian, A., Nah, S., Niverty, S., Page, D., Paschalidou, D., Patel, Z., Pavao, L., Ramezanali, M., Reda, F., Ren, X., Sabavat, V.R.N., Schmerling, E., Shi, S., Stefaniak, B., Tang, S., Tchapmi, L., Tredak, P., Tseng, W.C., Varghese, J., Wang, H., Wang, H., Wang, H., Wang, T.C., Wei, F., Wei, X., Wu, J.Z., Xu, J., Yang, W., Yen-Chen, L., Zeng, X., Zeng, Y., Zhang, J., Zhang, Q., Zhang, Y., Zhao, Q., Zolkowski, A.: Cosmos world foundation model platform for physical ai (2025), \url{https://arxiv.org/abs/2501.03575}

\bibitem{ouza2017simple}
Ouza, M., Ulrich, M., Yang, B.: A simple radar simulation tool for 3d objects based on blender. In: IRS (2017)

\bibitem{pan2024ratrack}
Pan, Z., Ding, F., Zhong, H., Lu, C.X.: Ratrack: moving object detection and tracking with 4d radar point cloud. In: ICRA (2024)

\bibitem{peebles2023scalable}
Peebles, W., Xie, S.: Scalable diffusion models with transformers. In: ICCV (2023)

\bibitem{piccinelli2025unidepthv2}
Piccinelli, L., Sakaridis, C., Yang, Y.H., Segu, M., Li, S., Abbeloos, W., Gool, L.V.: {U}ni{D}epth{V2}: Universal monocular metric depth estimation made simpler (2025), \url{https://arxiv.org/abs/2502.20110}

\bibitem{rafidashti2025neuradar}
Rafidashti, M., Lan, J., Fatemi, M., Fu, J., Hammarstrand, L., Svensson, L.: Neuradar: Neural radiance fields for automotive radar point clouds. In: CVPR (2025)

\bibitem{ran2024towards}
Ran, H., Guizilini, V., Wang, Y.: Towards realistic scene generation with lidar diffusion models. In: CVPR (2024)

\bibitem{10896467}
Rangaraj, P.A., Alkanat, T., Pandharipande, A.: Raids: Radar range-azimuth map estimation from image, depth, and semantic descriptions. IEEE Sensors Journal  (2025)

\bibitem{RemcomWaveFarer}
{Remcom Inc.}: Wavefarer® automotive radar software (2024)

\bibitem{ren2025cosmos}
Ren, X., Lu, Y., Cao, T., Gao, R., Huang, S., Sabour, A., Shen, T., Pfaff, T., Wu, J.Z., Chen, R., et~al.: Cosmos-drive-dreams: Scalable synthetic driving data generation with world foundation models. arXiv preprint arXiv:2506.09042  (2025)

\bibitem{rombach2021highresolution}
Rombach, R., Blattmann, A., Lorenz, D., Esser, P., Ommer, B.: High-resolution image synthesis with latent diffusion models (2021)

\bibitem{russell2025gaia}
Russell, L., Hu, A., Bertoni, L., Fedoseev, G., Shotton, J., Arani, E., Corrado, G.: Gaia-2: A controllable multi-view generative world model for autonomous driving. arXiv preprint arXiv:2503.20523  (2025)

\bibitem{sallab2019lidar}
Sallab, A.E., Sobh, I., Zahran, M., Essam, N.: Lidar sensor modeling and data augmentation with gans for autonomous driving. arXiv preprint arXiv:1905.07290  (2019)

\bibitem{saunders2024baseboostdepth}
Saunders, K., Manso, L.J., Vogiatzis, G.: Baseboostdepth: Exploiting larger baselines for self-supervised monocular depth estimation. arXiv preprint arXiv:2407.20437  (2024)

\bibitem{scharf1991statistical}
Scharf, L.L., Demeure, C.: Statistical signal processing: detection, estimation, and time series analysis. Prentice Hall (1991)

\bibitem{schoffmann2021virtual}
Sch{\"o}ffmann, C., Ubezio, B., B{\"o}hm, C., M{\"u}hlbacher-Karrer, S., Zangl, H.: Virtual radar: Real-time millimeter-wave radar sensor simulation for perception-driven robotics. IEEE RA-L  (2021)

\bibitem{schussler2021realistic}
Sch{\"u}{\ss}ler, C., Hoffmann, M., Br{\"a}unig, J., Ullmann, I., Ebelt, R., Vossiek, M.: A realistic radar ray tracing simulator for large mimo-arrays in automotive environments. IEEE Journal of Microwaves  (2021)

\bibitem{sligar2020machine}
Sligar, A.P.: Machine learning-based radar perception for autonomous vehicles using full physics simulation. IEEE Access  (2020)

\bibitem{song2025simulating}
Song, P., Song, D., Yang, Y., Lan, E., Liu, J.: Simulating automotive radar with lidar and camera inputs. arXiv preprint arXiv:2503.08068  (2025)

\bibitem{stetco2020radar}
Stetco, C., Ubezio, B., M{\"u}hlbacher-Karrer, S., Zangl, H.: Radar sensors in collaborative robotics: Fast simulation and experimental validation. In: IEEE ICRA (2020)

\bibitem{stowell2008investigation}
Stowell, M.L., Fasenfest, B.J., White, D.A.: Investigation of radar propagation in buildings: A 10-billion element cartesian-mesh fetd simulation. IEEE transactions on antennas and propagation  (2008)

\bibitem{thieling2020scalable}
Thieling, J., Frese, S., Ro{\ss}mann, J.: Scalable and physical radar sensor simulation for interacting digital twins. IEEE Sensors Journal  (2020)

\bibitem{tibshirani1996regression}
Tibshirani, R.: Regression shrinkage and selection via the lasso. Journal of the Royal Statistical Society Series B: Statistical Methodology  (1996)

\bibitem{topak2011system}
Topak, A.E., Hasch, J., Zwick, T.: A system simulation of a 77 ghz phased array radar sensor. In: IRS (2011)

\bibitem{tyszkiewicz2023gecco}
Tyszkiewicz, M.J., Fua, P., Trulls, E.: Gecco: Geometrically-conditioned point diffusion models. In: ICCV (2023)

\bibitem{vahdat2022lion}
Vahdat, A., Williams, F., Gojcic, Z., Litany, O., Fidler, S., Kreis, K., et~al.: Lion: Latent point diffusion models for 3d shape generation. NeurIPS  (2022)

\bibitem{weston2021there}
Weston, R., Jones, O.P., Posner, I.: There and back again: Learning to simulate radar data for real-world applications. In: IEEE ICRA (2021)

\bibitem{wheeler2017deep}
Wheeler, T.A., Holder, M., Winner, H., Kochenderfer, M.J.: Deep stochastic radar models. In: IEEE IV (2017)

\bibitem{wu2025chronoedit}
Wu, J.Z., Ren, X., Shen, T., Cao, T., He, K., Lu, Y., Gao, R., Xie, E., Lan, S., Alvarez, J.M., Gao, J., Fidler, S., Wang, Z., Ling, H.: Chronoedit: Towards temporal reasoning for image editing and world simulation. arXiv preprint arXiv:2510.04290  (2025)

\bibitem{wu2024text2lidar}
Wu, Y., Zhang, K., Qian, J., Xie, J., Yang, J.: Text2lidar: Text-guided lidar point cloud generation via equirectangular transformer. In: ECCV (2024)

\bibitem{wu2023sketch}
Wu, Z., Wang, Y., Feng, M., Xie, H., Mian, A.: Sketch and text guided diffusion model for colored point cloud generation. In: ICCV (2023)

\bibitem{xiao2025simulate}
Xiao, W., Huang, H., Zhong, C., Lin, Y., Wang, N., Chen, X., Chen, Z., Zhang, S., Yang, S., Merriaux, P., et~al.: Simulate any radar: Attribute-controllable radar simulation via waveform parameter embedding. arXiv preprint arXiv:2506.03134  (2025)

\bibitem{xie2024sana}
Xie, E., Chen, J., Chen, J., Cai, H., Tang, H., Lin, Y., Zhang, Z., Li, M., Zhu, L., Lu, Y., et~al.: Sana: Efficient high-resolution image synthesis with linear diffusion transformers. arXiv preprint arXiv:2410.10629  (2024)

\bibitem{xiong2023learning}
Xiong, Y., Ma, W.C., Wang, J., Urtasun, R.: Learning compact representations for lidar completion and generation. In: CVPR (2023)

\bibitem{yang2019pointflow}
Yang, G., Huang, X., Hao, Z., Liu, M.Y., Belongie, S., Hariharan, B.: Pointflow: 3d point cloud generation with continuous normalizing flows. In: ICCV (2019)

\bibitem{yun2015ray}
Yun, Z., Iskander, M.F.: Ray tracing for radio propagation modeling: Principles and applications. IEEE access  (2015)

\bibitem{zamorski2020adversarial}
Zamorski, M., Zi{\k{e}}ba, M., Klukowski, P., Nowak, R., Kurach, K., Stokowiec, W., Trzci{\'n}ski, T.: Adversarial autoencoders for compact representations of 3d point clouds. Computer Vision and Image Understanding  (2020)

\bibitem{zhang2025ufm}
Zhang, Y., Keetha, N., Lyu, C., Jhamb, B., Chen, Y., Qiu, Y., Karhade, J., Jha, S., Hu, Y., Ramanan, D., Scherer, S., Wang, W.: Ufm: A simple path towards unified dense correspondence with flow. In: arXiv (2025)

\bibitem{zhou20213d}
Zhou, L., Du, Y., Wu, J.: 3d shape generation and completion through point-voxel diffusion. In: ICCV (2021)

\bibitem{zyrianov2022learning}
Zyrianov, V., Zhu, X., Wang, S.: Learning to generate realistic lidar point clouds. In: ECCV (2022)

\end{thebibliography}
